\documentclass[accepted]{uai2026} 

\usepackage[american]{babel}

\usepackage{natbib} 
\usepackage{mathtools} 
\usepackage{booktabs} 
\usepackage{tikz} 

\usepackage{graphicx}
\usepackage{subcaption}
\usepackage{float}
\usepackage{amsmath}
\usepackage{amsfonts}
\usepackage{amssymb}

\title{Causal Discovery on Irregular Time Series}

\author[1,2]{Martim Penim}
\author[1,3]{Ricardo Ribeiro Pereira}
\author[1]{Jacopo Bono}
\author[1]{\\Hugo Ferreira}
\author[2,4]{Mário A.T. Figueiredo}
\author[1]{Pedro Bizarro}
\affil[1]{%
    Feedzai\\
    Portugal
}
\affil[2]{%
    Instituto Superior Técnico, Universidade de Lisboa\\
    Portugal
}
\affil[3]{%
    DCC, Faculdade de Ciêncidas da Universidade do Porto\\
    Portugal
}
\affil[4]{%
    Instituto de Telecomunicações\\
    Portugal
}

\begin{document}
\maketitle

\begin{abstract}
Causal discovery methods have shown strong performance in temporal systems, but they typically rely on regular and discrete lag structures, limiting their applicability to regularly sampled data.
However, many real-world tasks require dealing with irregularly sampled streams of events, such as sensor streams, healthcare data, and financial transactions.
In this work, we propose an extension of PCMCI+, a state-of-the-art method for causal discovery on regular multivariate time series, to allow for handling irregular time series.
Instead of modelling causal relations through fixed-lag dependencies, our method aggregates causal influence over predefined temporal windows.
We evaluate our method on synthetic irregular event streams with known causal structures under different signal-to-noise ratios, showing that it consistently recovers the underlying causal graph and substantially outperforms the standard PCMCI+ on irregularly sampled data.
\end{abstract}

\section{Introduction}
\label{sec:intro}

Many real-world decision-making systems operate on sequential data, generated at irregular times.
Financial transactions, healthcare events, user interactions, and sensor streams are naturally event-driven, meaning that observations do not occur at fixed temporal intervals.
Understanding the causal structure underlying such data is essential for robust prediction, risk assessment, anomaly detection, and ultimately informed decision-making under uncertainty.


Causal discovery methods for time series have achieved significant progress in recent years, particularly through frameworks based on conditional independence tests and graphical models.
Among these, PCMCI+ \citep{runge2020discovering} has emerged as a powerful approach to recover causal relations in high-dimensional temporal systems, due to its scalability, robustness to autocorrelation, and support for nonlinear dependencies. However, PCMCI+ implicitly assumes regularly sampled observations, with temporal dependencies over discrete lags, which hampers its use in real-world problems where events occur irregularly.

In this work, we propose a time-aware extension of PCMCI+ for irregularly sampled time series.
Our approach preserves the original conditional-independence-based framework, while redefining how temporal dependencies are constructed under irregular event timing.
As opposed to finding causal relations across fixed discrete lags (``exactly 5 events ago''), our method therefore aims to discover causal relations across time (``around 5 minutes ago''). 

\section{Related Work}
\label{sec:relatedwork}

Early work on causal discovery in time series focused on approaches such as \textit{Granger causality} \citep{Granger1969} and \textit{vector autoregressive} models \citep{Asteriou}, which evaluate whether past values of one variable improve the prediction of another.
While influential, these methods are often limited to linear or low-dimensional settings.

Constraint-based methods such as PC~\citep{spirtes1991algorithm} and FCI~\citep{spirtes2000causation}  infer causal structure through conditional independencies represented as graphical models.
However, they were not specifically designed for temporal data and can become computationally expensive or statistically unstable in multivariate time series with strong autocorrelation.

To address these limitations, the PCMCI~\citep{runge2019detecting} and PCMCI+~\citep{runge2020discovering} methods were specifically designed for causal discovery in large multivariate time series.
PCMCI+ combines a condition selection stage with conditional independence testing, substantially reducing dimensionality while remaining robust to autocorrelation and indirect dependencies \citep{miersch2025evaluating, nakamura2025cloud}.
Unlike classical Granger-based approaches, PCMCI+ provides a framework that can be extended to nonlinear dependencies through suitable independence tests, while explicitly distinguishing between lagged and contemporaneous causal relations.
Despite these advances, PCMCI and PCMCI+ assume regularly sampled observations, which becomes problematic in domains where observations occur at irregular times.

A common approach to handle irregular sampling is to transform observations onto a regular grid through interpolation or aggregation, enabling the use of standard time-series methods \citep{shukla2020survey}. 
However, resampling can distort temporal dynamics, blur causal delays, and introduce artificial dependencies, especially when event frequencies vary over time.

An alternative line of work models irregular observations directly in continuous time.
Point processes, including Hawkes processes, naturally capture asynchronous interactions, while related methods extend continuous-time conditional independence testing to causal discovery in marked point processes \citep{didelez2008graphical, christgau2023nonparametric}.
However, these approaches differ substantially from the PCMCI+ framework, as they model continuous-time event intensities rather than lagged dependencies between variables.

More recent approaches use neural continuous-time representations to jointly learn latent dynamics and causal structure from irregular observations \citep{bellot2021neural, cheng2023cuts, cheng2024cuts+}.
While flexible for modeling nonlinear dependencies, these methods are typically optimization-heavy, less interpretable, and rely on fundamentally different inference mechanisms from the framework used in PCMCI+.

Our work aims to extend the PCMCI+ core causal-discovery mechanism to irregularly sampled data while preserving its interpretability, scalability, and conditional-independence-based methodology.
In this sense, the proposed methods can be viewed as a bridge between the PCMCI framework and the broader literature on irregular temporal causal discovery.

\section{Methods}
\label{methods}

PCMCI+ is a causal discovery algorithm designed for regularly sampled time series, where observations are assumed to occur at equally spaced intervals.
More specifically, PCMCI+ repeatedly evaluates relations of the form $X^i_{t-\tau} \perp X^j_t \mid S$, where $i$ and $j$ are indices of variables, $t$ is the time index, $\tau$ denotes a temporal lag measured in discrete event steps and $S$ is a conditioning set composed of other lagged variables.
For each valid time index $t$, the algorithm constructs triplets of the form $(X^i_{t-\tau}, X^j_t, S)$, which are then passed to a conditional independence test.

In regularly sampled data, this procedure is straightforward because each lag corresponds to a unique past observation.
However, this structure breaks down if observations are not evenly spaced in time, because equal index differences do not correspond to equal temporal delays and the notion of a discrete lag becomes ambiguous.
For example, a pair of consecutive events ($\tau=1$) can be a few seconds apart, while another consecutive pair in the same sequence might be separated by hours.
Applying the standard PCMCI+ method directly would therefore produce temporally inconsistent samples and potentially misleading causal relations.

Rather than modifying the conditional independence machinery itself, our approach reformulates how the triplets are constructed.
The central idea is to replace index-based lag alignment with time-aware pairing mechanisms that operate directly on timestamps.
In this way, the original structure of PCMCI+ can largely be preserved while extending its applicability to irregularly sampled data.
Formally, given two observations occurring at times $t'$ and $t$, we seek pairs satisfying $t - t' \approx \Delta$, for some target temporal delay $\Delta$.
This way, the challenge is transformed from one of modifying causal inference into one of constructing meaningful temporal neighbourhoods under irregular sampling.
To address it, we introduce two options for extending PCMCI+: a rectangular-window variant and a Gaussian-window variant.

\subsection{Rectangular-Window PCMCI+}

The first variant replaces discrete event step lags with time-difference windows.
Instead of pairing observations that are exactly $\tau$ events apart, we pair each event with the observations that fall within a predefined temporal interval around the target delay.

Given a current event occurring at time $t$, and a lag index $k$, we define a temporal window of width $\Delta$ as
\[
[t - k\Delta,\; t - (k-1)\Delta).
\]
Any observation whose timestamp falls inside this interval is treated as a valid realization of lag $k$.
Consequently, each lag is no longer associated with a single past observation, but with a set of temporally nearby events.
This also implies that a single event may contribute to multiple constructed triplets, unlike the standard PCMCI+ where each event contributes to exactly one triplet.

This modification introduces a continuous notion of temporal proximity while preserving the overall causal discovery pipeline of PCMCI+.
The conditional independence tests remain unchanged; only the mechanism used to construct aligned triplets is adapted to account for irregular timing.

To further improve robustness under various event distributions, we additionally consider overlapping windows.
In this setting, consecutive lag windows partially intersect, allowing observations near window boundaries to contribute to multiple lag representations. 
This reduces sensitivity to arbitrary discretization effects, making the representation less dependent on the exact placement of window edges.

\subsection{Gaussian-Window PCMCI+}

While the previous variant introduces flexibility through discrete temporal neighbourhoods, it still imposes hard boundaries between observations. 
To soften this behaviour, we propose a second variant based on continuous temporal weighting.
Instead of assigning observations to lag windows in a binary manner, the Gaussian-window variant weights observations according to their temporal distance from the target delay. 
Observations closer to the centre of the window contribute more strongly than distant observations.

Let $\delta = t - t'$ denote the temporal distance between two observations.
We define a weighting function
\[
w(\delta ; k, \mu, \sigma)
=
\exp\!\left(-(\delta - k\mu)^2 / (2\sigma^2)\right),
\]
where $k\mu$ is the target delay and $\sigma$ controls the decay of influence.
This way, observations with temporal distances close to $k\mu$ receive higher weights, whereas observations that are further from $k\mu$ are smoothly down-weighted.

This approach softens the discontinuities introduced by fixed windows and provides a softer representation of temporal influence.
It allows the algorithm to exploit all available observations while still respecting temporal structure.


\section{Synthetic data generation}
To evaluate the proposed methods in a controlled setting, we generated synthetic irregular time series with known causal structure.
The simulated data consists of two components: a causal event stream and an independent noise stream.

Timestamps for causal events were generated over a fixed observation horizon using random inter-event time intervals $\Delta_m\sim\mathcal{N}(\mu_c,\sigma_c^2)$, centred around $\mu_c$ and with $\sigma_c$ controlling the degree of temporal irregularity and chosen such that $\mathbb{P}(\Delta_m < 0) \simeq 0$.
Then, their feature values were generated according to a first-order autoregressive process defined over event indices.
With $A$ denoting the adjacency matrix of the ground-truth causal graph, for each event index $m$, features were generated as
\begin{equation}
\label{eq:var}
X_m^{(j)} = \varepsilon_m^{(j)} + \sum_{i} A_{ji} X_{m-1}^{(i)},
\end{equation}
where $\varepsilon_m^{(j)}$ is Gaussian noise.
This construction isolates the effect of irregular temporal sampling from the underlying causal mechanism, ensuring that differences in performance arise from temporal handling rather than changes in the data-generating process.

To emulate realistic event streams containing irrelevant observations, additional noise events were generated independently using a Poisson process and merged with the causal stream.
Their features consist of random Gaussian noise and are not causally related.

The experiments were conducted on synthetic irregular event streams generated over a simulation horizon of 500 days, with causal event timestamps produced using $\mu_c = 6$ hours and standard deviation $\sigma_c = 0.25$ hours.
We provide an exploratory analysis of this data in Appendix~\ref{sec:appendix_times}.

The underlying system contains 13 variables organized into canonical causal structures, including chains, forks, colliders, and diamond patterns, while 2 additional variables act as independent noise variables.
All causal relationships were generated through Equation~\ref{eq:var} with edge weights (nonzero elements of the adjacency matrix) fixed at $0.5$, and Gaussian noise $\epsilon_m^{(j)} \sim \mathcal{N}(0,1)$ (Appendix~\ref{sec:appendix_features}).

To evaluate robustness against contamination, several causal-to-noise event ratios were considered, namely 1:0, 2:1, 1:1, 1:2, 1:3, and 1:4. 
Increasing the proportion of noise events creates progressively more challenging conditions for causal discovery. 

\section{Results}
\label{sec:results}

For all configurations, causal discovery was performed using PCMCI+ with the \textit{partial correlation} conditional independence test and a significance level of 0.01. For standard PCMCI+, the maximum lag was set to $\tau_{\max}=10$ to account for large effective delays introduced by irregular sampling and noise events.

Four experiments were designed to evaluate how temporal alignment and window overlap affect causal discovery in irregular event streams.
In all experiments, the Gaussian windows are set to overlap with a corresponding rectangular windows.
The only difference is that the latter weights all events evenly, while the former gives more weight to events in the centre of the Gaussian window (Appendix~\ref{sec:appendix_windows}).

\textbf{Experiment 1:}
We start by considering an ideal aligned setting in which the temporal representations are centred around the true causal delay of 6 hours.
The intervals used were $[-4,0),\; [-8,-4),\; [-12,-8)$ hours, such that the average delay of the causal stream lies near the centre of the second window. 

\textbf{Experiment 2:}
We then evaluate the effect of temporal misalignment.
The window configuration is shifted to $[-6,-2),\; [-10,-6),\; [-14,-10)$ hours, placing the true 6-hour delay directly on a boundary between neighbouring temporal regions. 

\textbf{Experiment 3:}
We introduce overlap between neighbouring temporal representations, resulting in the following intervals $[-4,0),\; [-6,-2),\; [-8,-4),\; ...,\; [-14,-10)$ hours. 

\textbf{Experiment 4:}
We increase the temporal resolution and reduce window width from 4 hours to 2 hours, producing the sequence $[-2,0),\; [-3,-1),\; ...,\; [-13,-11)$ hours. 

\vspace{3mm}

To assess the results, we use the \textit{Structural Hamming Distance} (SHD), while recall and precision are provided in Appendix~\ref{sec:appendixA}.
To avoid relying on a single edge-selection cutoff that could unfairly favour particular methods, we additionally evaluate all detected edges across p-value thresholds below 0.01 and report the threshold yielding the best SHD for each method and noise level.
As such, the reported performance for each method represents the upper-bound of the optimal p-value optimization strategy.

\begin{figure}[H]
    \centering
    \begin{subfigure}{0.49\columnwidth}
        \centering
        \includegraphics[width=\linewidth]{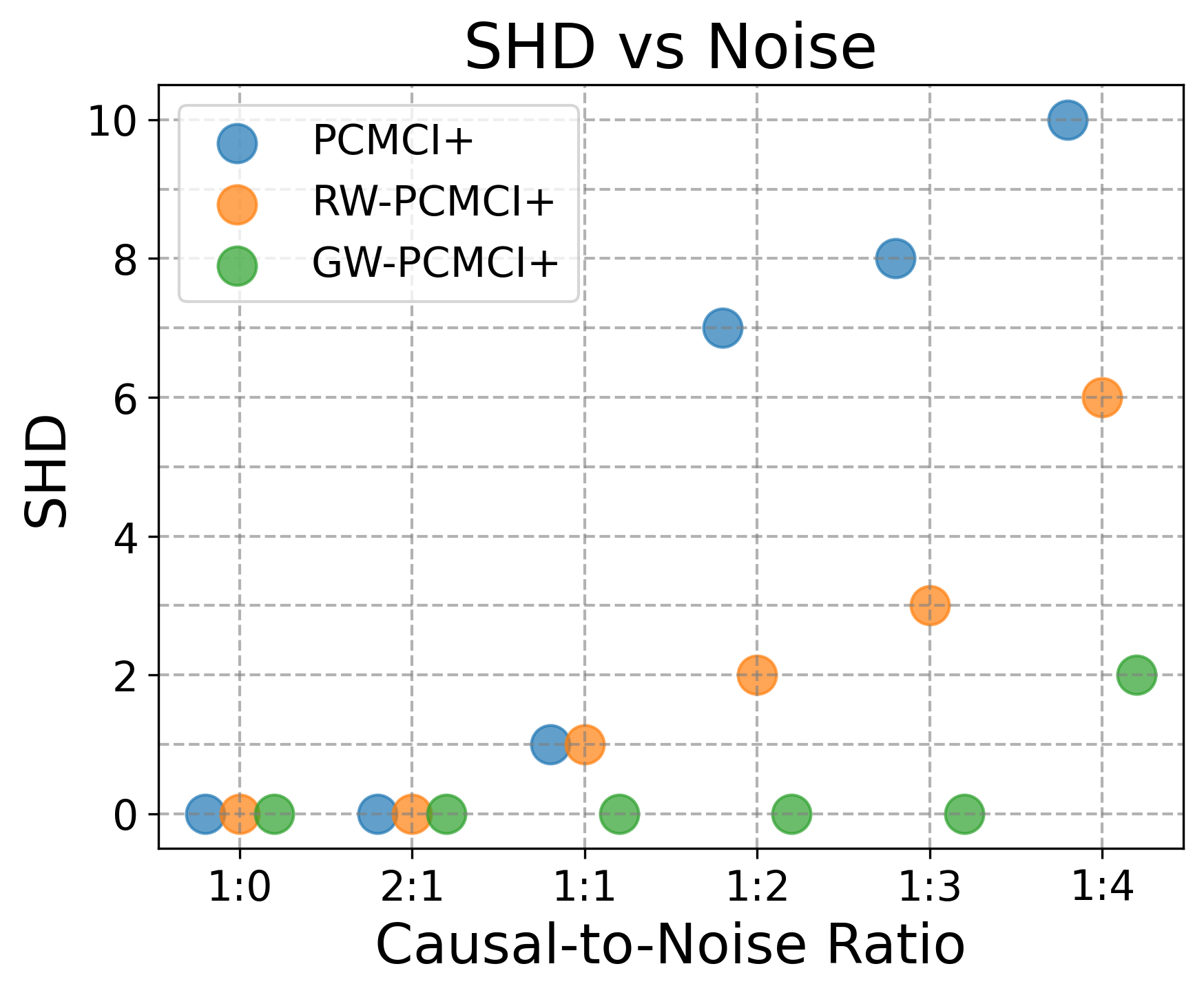}
    \end{subfigure}
    \hfill
    \begin{subfigure}{0.49\columnwidth}
        \centering
        \includegraphics[width=\linewidth]{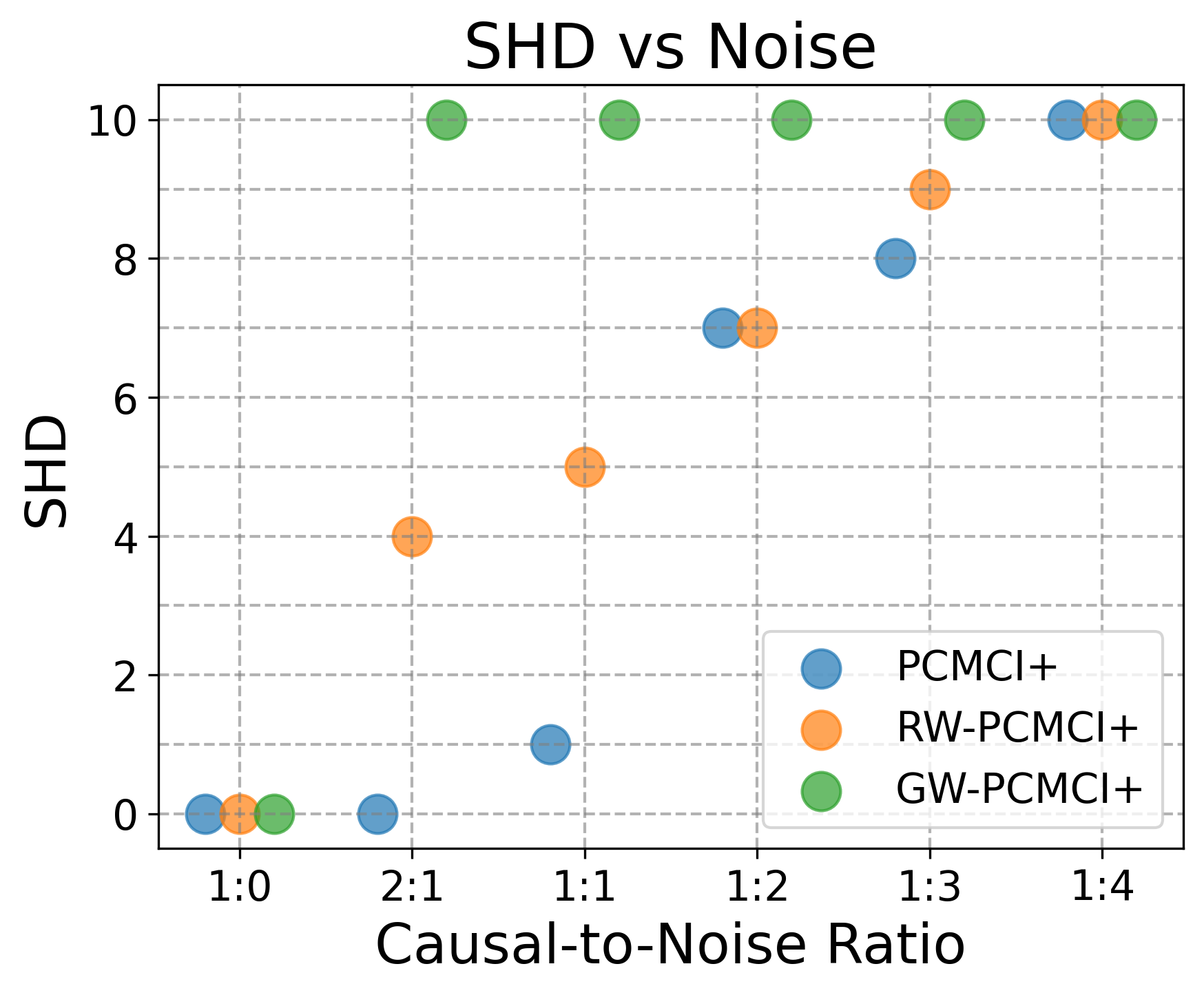}
    \end{subfigure}

    \vspace{4mm}

    \begin{subfigure}{0.49\columnwidth}
        \centering
        \includegraphics[width=\linewidth]{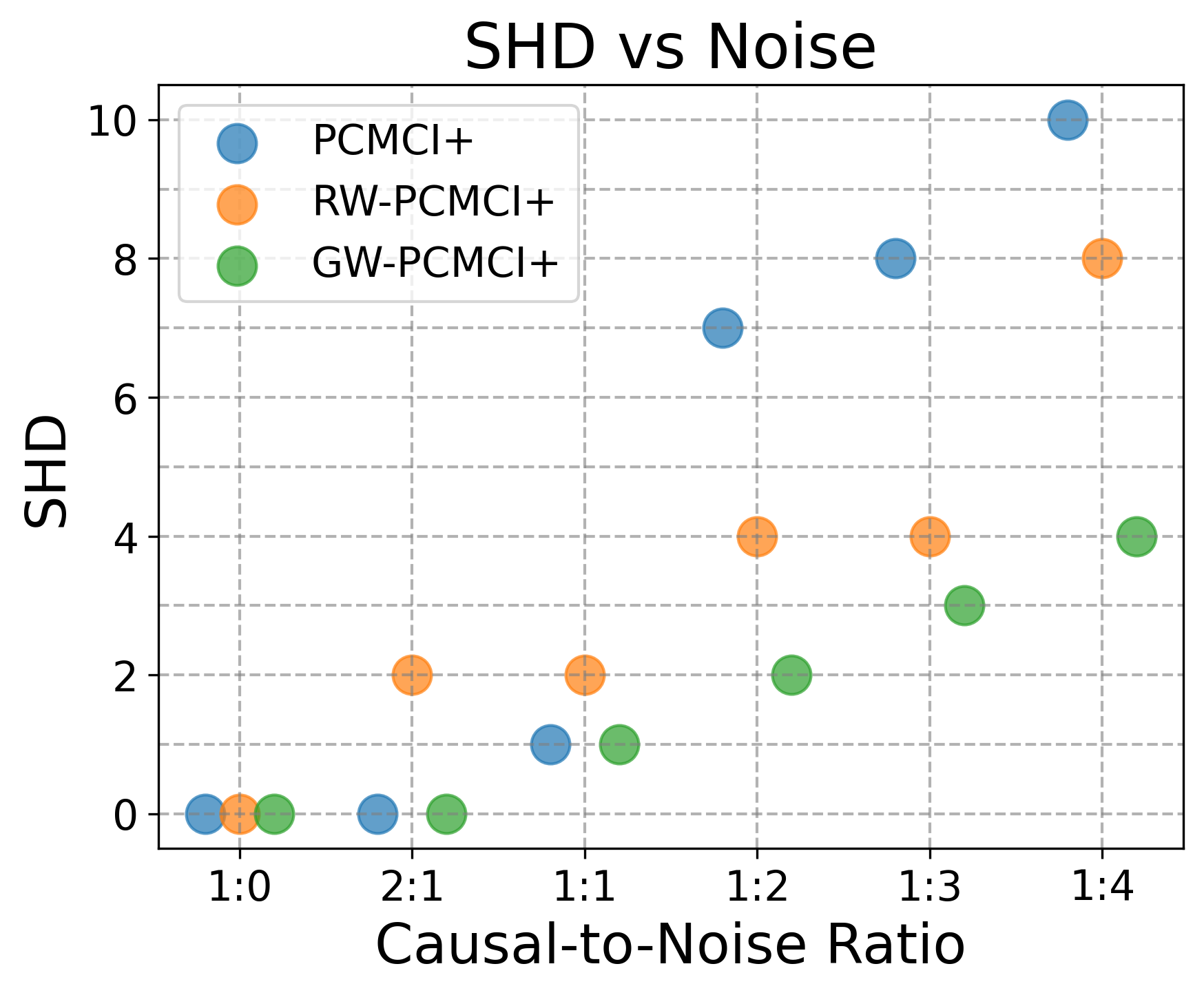}
    \end{subfigure}
    \hfill
    \begin{subfigure}{0.49\columnwidth}
        \centering
        \includegraphics[width=\linewidth]{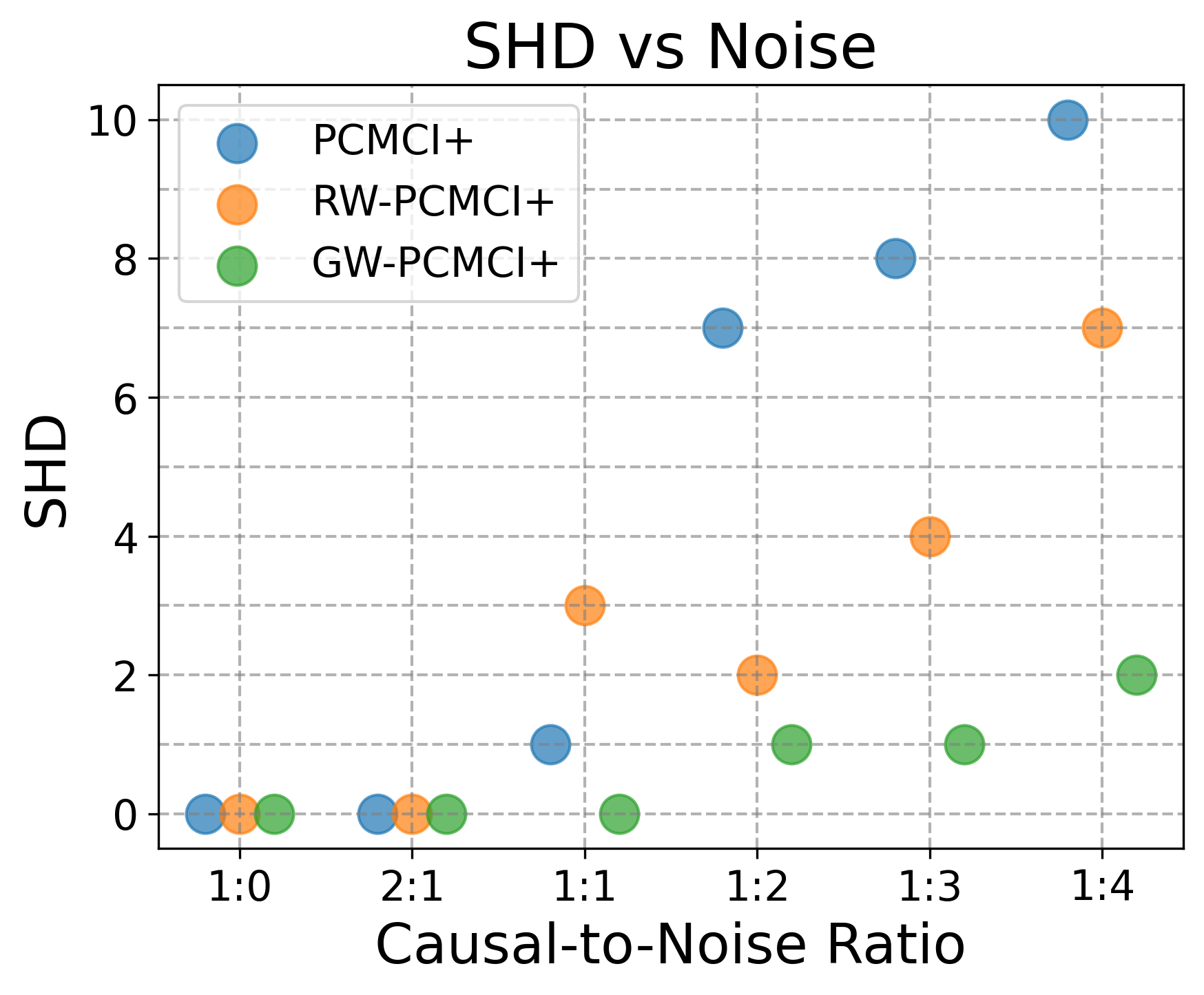}
    \end{subfigure}

    \caption{SHD for experiments 1 (top-left), 2 (top-right), 3 (bottom-left), and 4 (bottom-right) across different causal-to-noise ratios.}
    \label{fig:SHD}
\end{figure}

The results in Figure~\ref{fig:SHD} show that both variants of the proposed extension outperform standard PCMCI+ in irregular event streams, especially in lower causal-to-noise ratio settings.

In the first experiment, where the temporal representations are correctly aligned with the true causal delay, both the rectangular and Gaussian windows achieve lower SHD across all noise levels.
When the noise level is low, all methods perform well as expected, because the sequence of causal events is essentially regular. However, as the proportion of noise events increases, the performance of standard PCMCI+ deteriorates rapidly, while both versions of the proposed method remain substantially more robust.
This indicates that explicitly incorporating temporal information is critical in irregular settings.

The second experiment demonstrates the importance of temporal alignment.
When the windows are not correctly aligned with the true causal delay, the performance decreases considerably compared to the aligned setting.
This degradation confirms that the quality of the temporal representation strongly influences causal recovery.
In particular, placing the true delay near boundaries between temporal windows makes the discovery problem significantly harder. 
This effect is especially pronounced for the Gaussian-window variant.
Because the true delays now fall near the valleys between adjacent Gaussian kernels, the corresponding observations receive very low weights and therefore contribute only weakly to the causal estimates.
In contrast, the rectangular-window variant is less affected, since all observations within a window are weighted equally regardless of their exact temporal position.

The third experiment shows that overlapping temporal representations effectively solve this issue.
By allowing multiple windows to cover the same temporal region, the performance of the methods gets closer to the one of aligned setting. 

The fourth experiment shows that increasing the temporal resolution by using smaller and denser windows allows the methods to localize causal dependencies more precisely, leading to more accurate recovery of the true causal graph.
The experiments suggest that overlap and fine-grained temporal representations are key ingredients for robust causal discovery in irregular time series. 

Among all approaches (excluding Experiment 2), the Gaussian-window variant consistently achieves the best performance, yielding low SHD values across all noise levels.

\section{Conclusion}
\label{sec:conclusion}

In this work, we proposed an extension of PCMCI+ for causal discovery in irregularly sampled time series.
We show that incorporating temporal information substantially improves causal discovery in irregular event streams.
While standard PCMCI+ rapidly deteriorates as regularity decreases, the proposed methods remain more robust.
Among all methods, the Gaussian-window variant achieved the best overall performance.

These results suggest that preserving the conditional-independence-based semantics of PCMCI+ while adapting the temporal representation provides an effective strategy for extending causal discovery to event-driven environments.

Future work should focus on evaluating the proposed methods on real-world irregular datasets, in domains such as financial transactions and healthcare event streams.
Additional directions include extending the framework to non-linear conditional independence tests and settings involving latent confounding, non-stationarity, or continuous-time causal dynamics.
Another promising direction is the exploration of more general temporal weighting kernels that unify the rectangular and Gaussian formulations considered in this work. Such approaches could provide a flexible continuum between hard temporal windows and smooth distance-based weighting, potentially improving robustness across different irregular sampling regimes.

\bibliography{uai2026-template}

\newpage

\onecolumn

\title{Causal Discovery on Irregular Time Series\\(Supplementary Material)}
\maketitle

\appendix

\section{Exploratory Data Analysis}

We provide an exploratory analysis of the synthetic irregular event streams used throughout the experiments. The objective of this analysis is to better characterize the temporal properties of the generated data, understand how increasing noise affects the event dynamics, and illustrate how the underlying causal structures manifest in the observed feature trajectories. 

\subsection{Interarrival Times}
\label{sec:appendix_times}
\begin{figure}[ht]
    \centering
    \begin{subfigure}{0.33\columnwidth}
        \centering
        \includegraphics[width=.95\linewidth]{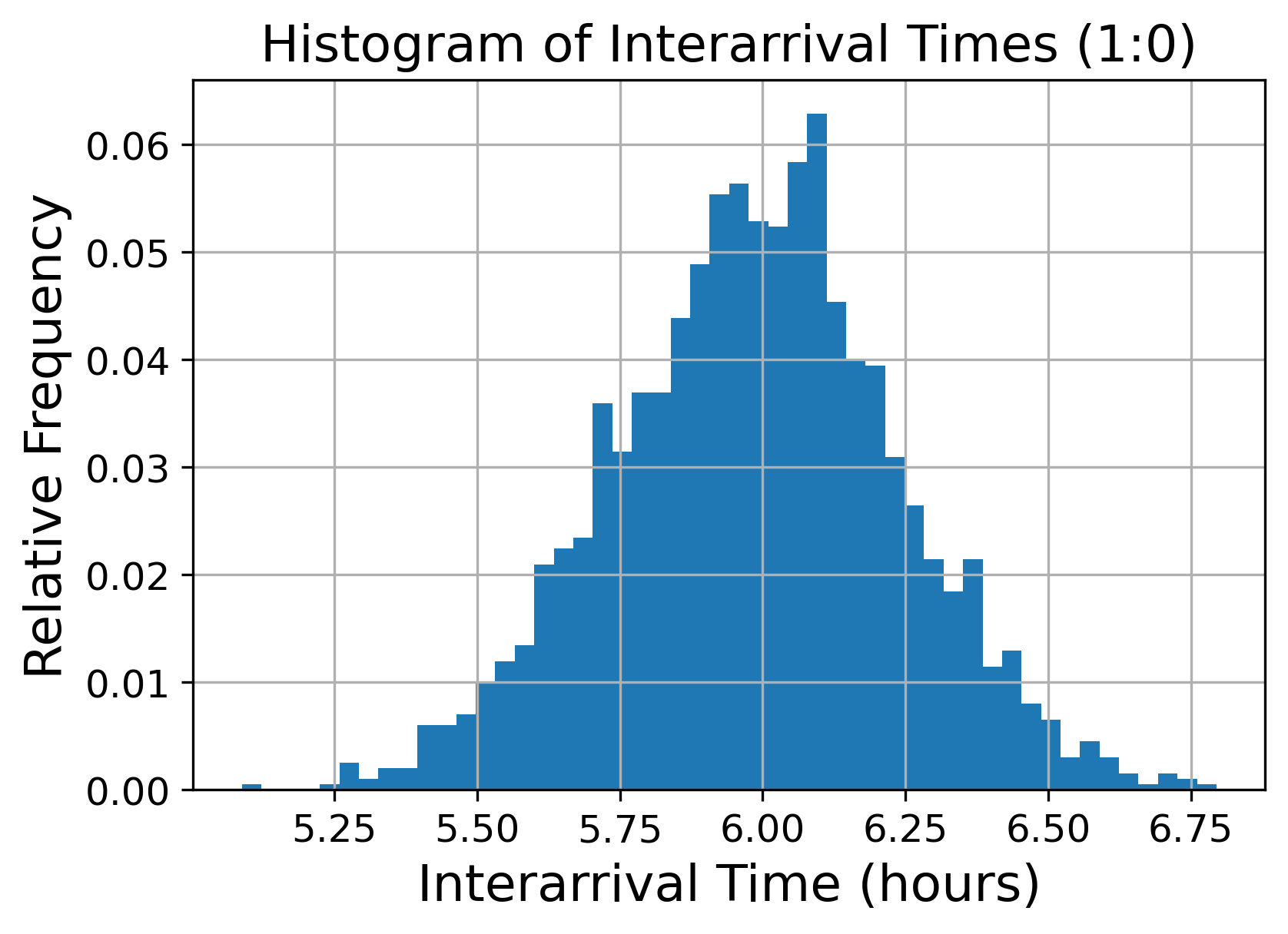}
    \end{subfigure}
    \hfill
    \begin{subfigure}{0.33\columnwidth}
        \centering
        \includegraphics[width=.95\linewidth]{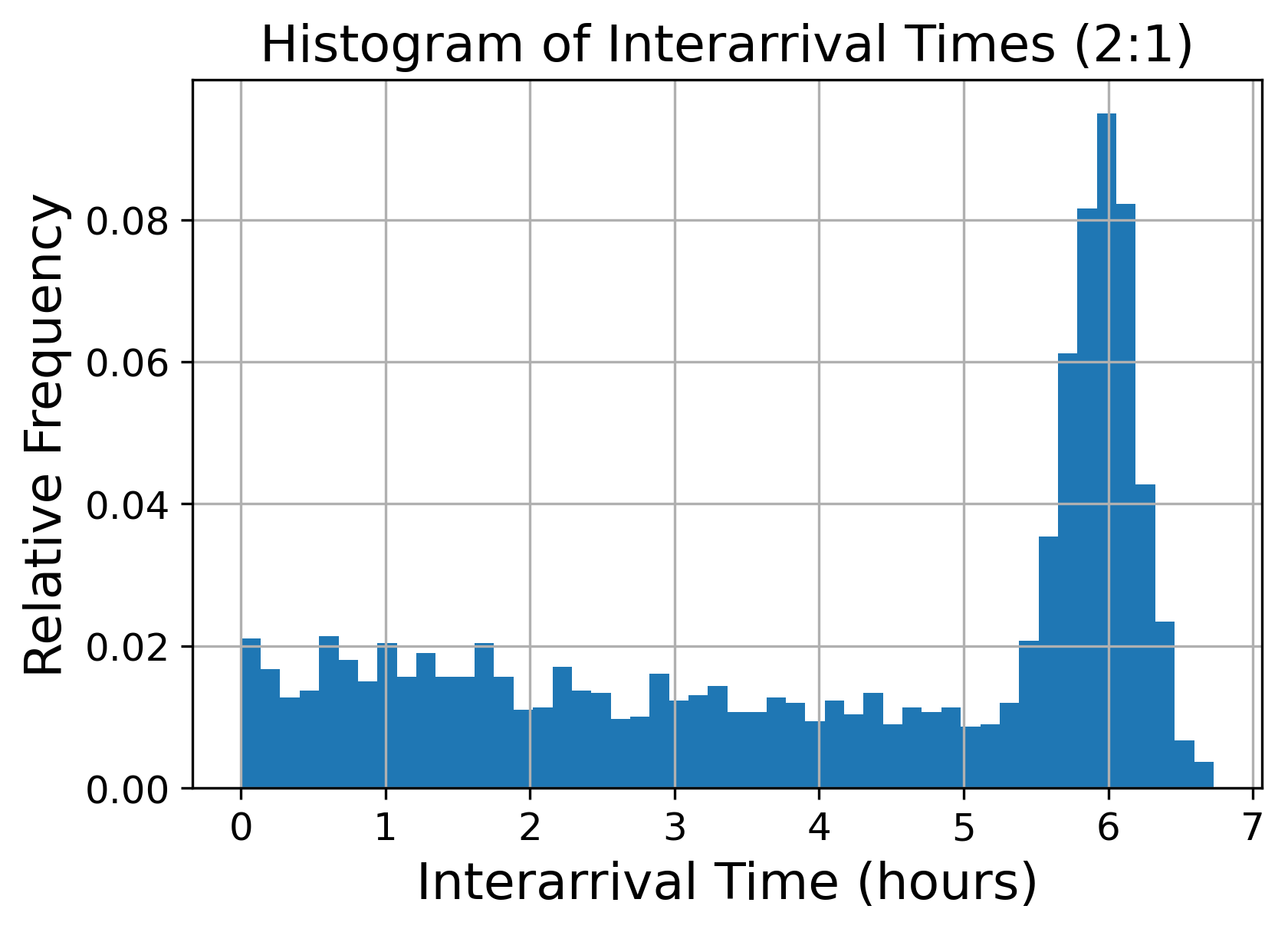}
    \end{subfigure}
    \begin{subfigure}{0.33\columnwidth}
        \centering
        \includegraphics[width=.95\linewidth]{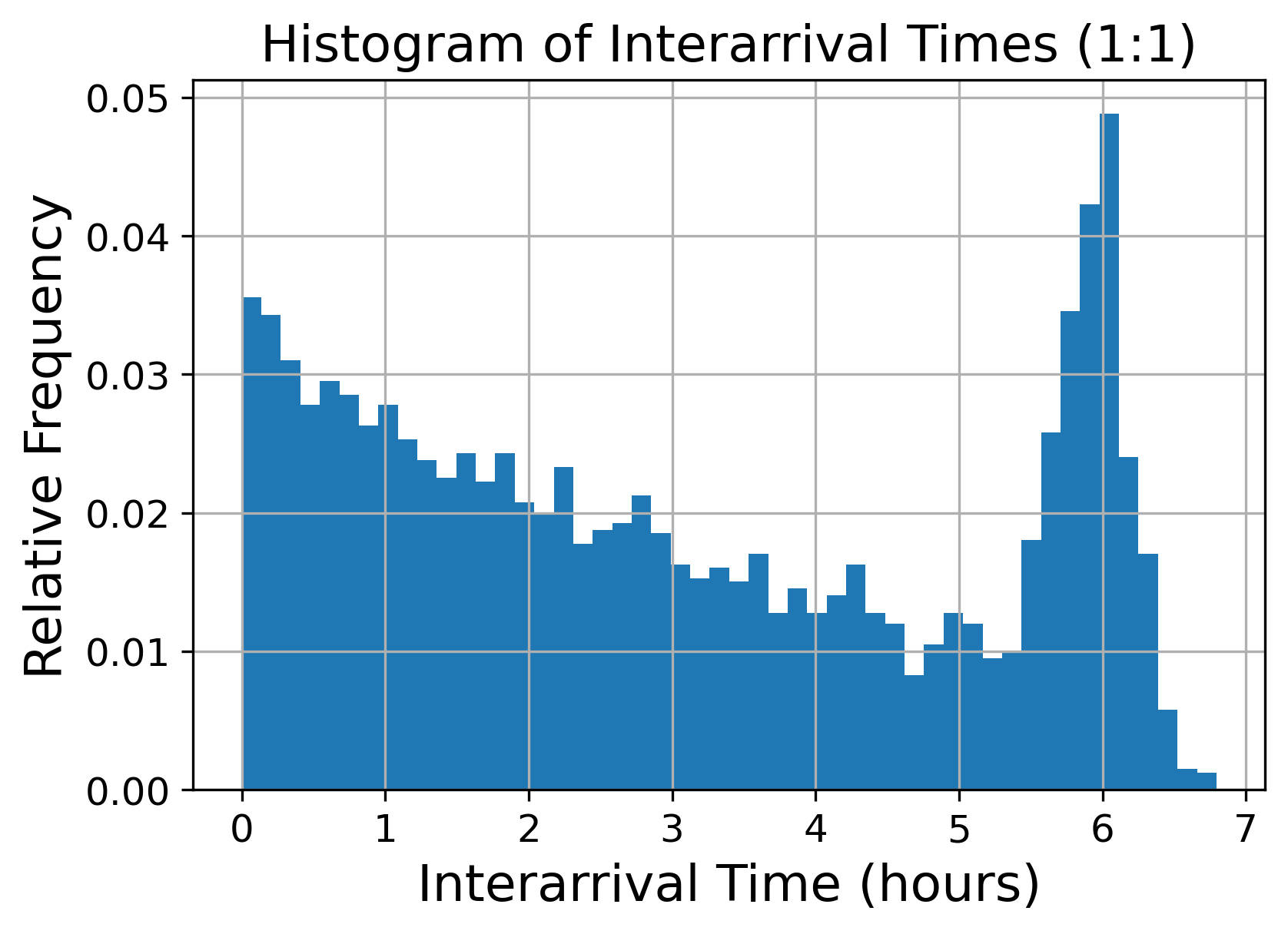}
    \end{subfigure}
    \hfill
    \begin{subfigure}{0.33\columnwidth}
        \centering
        \includegraphics[width=.95\linewidth]{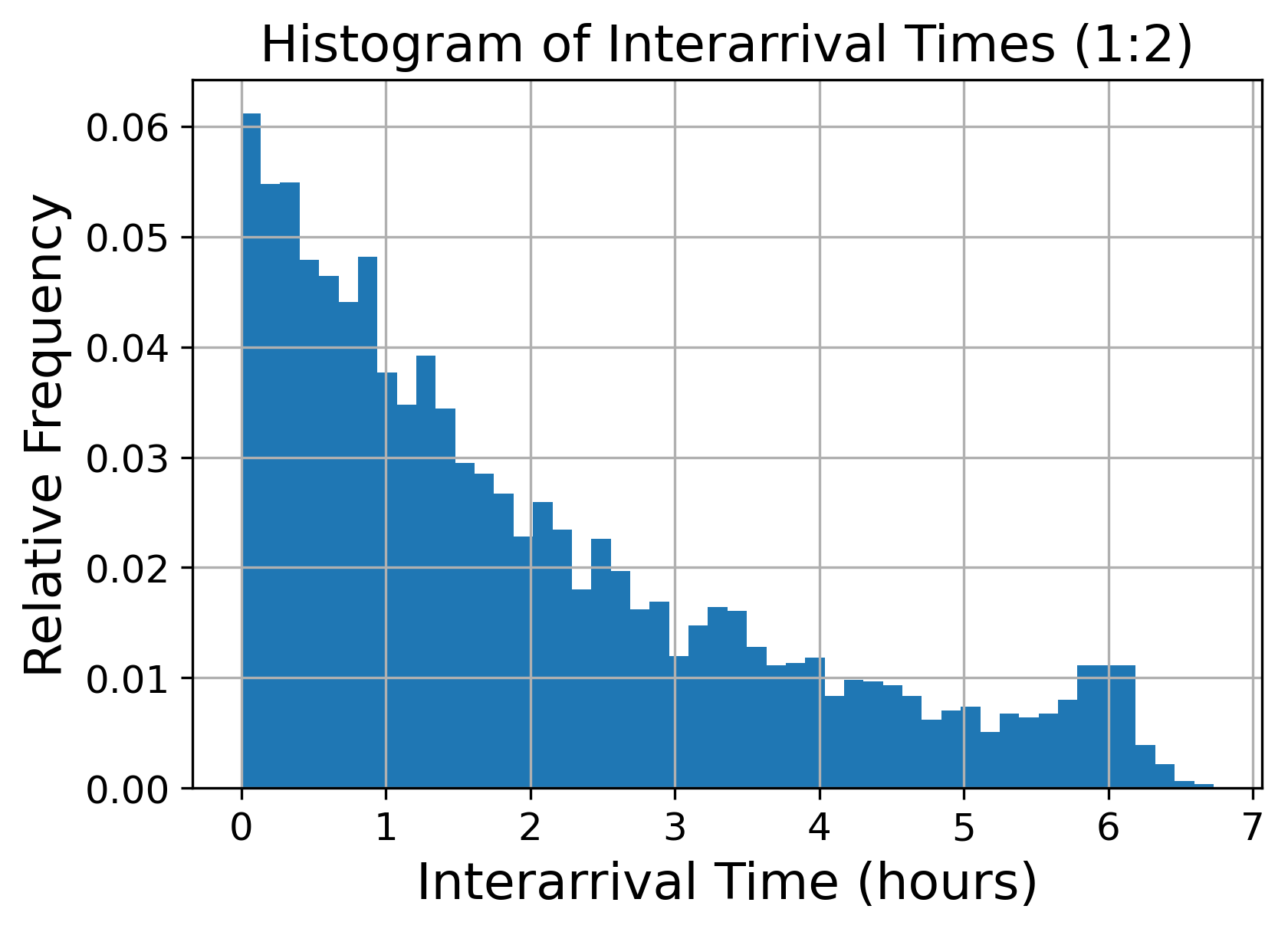}
    \end{subfigure}
        \hfill
    \begin{subfigure}{0.33\columnwidth}
        \centering
        \includegraphics[width=.95\linewidth]{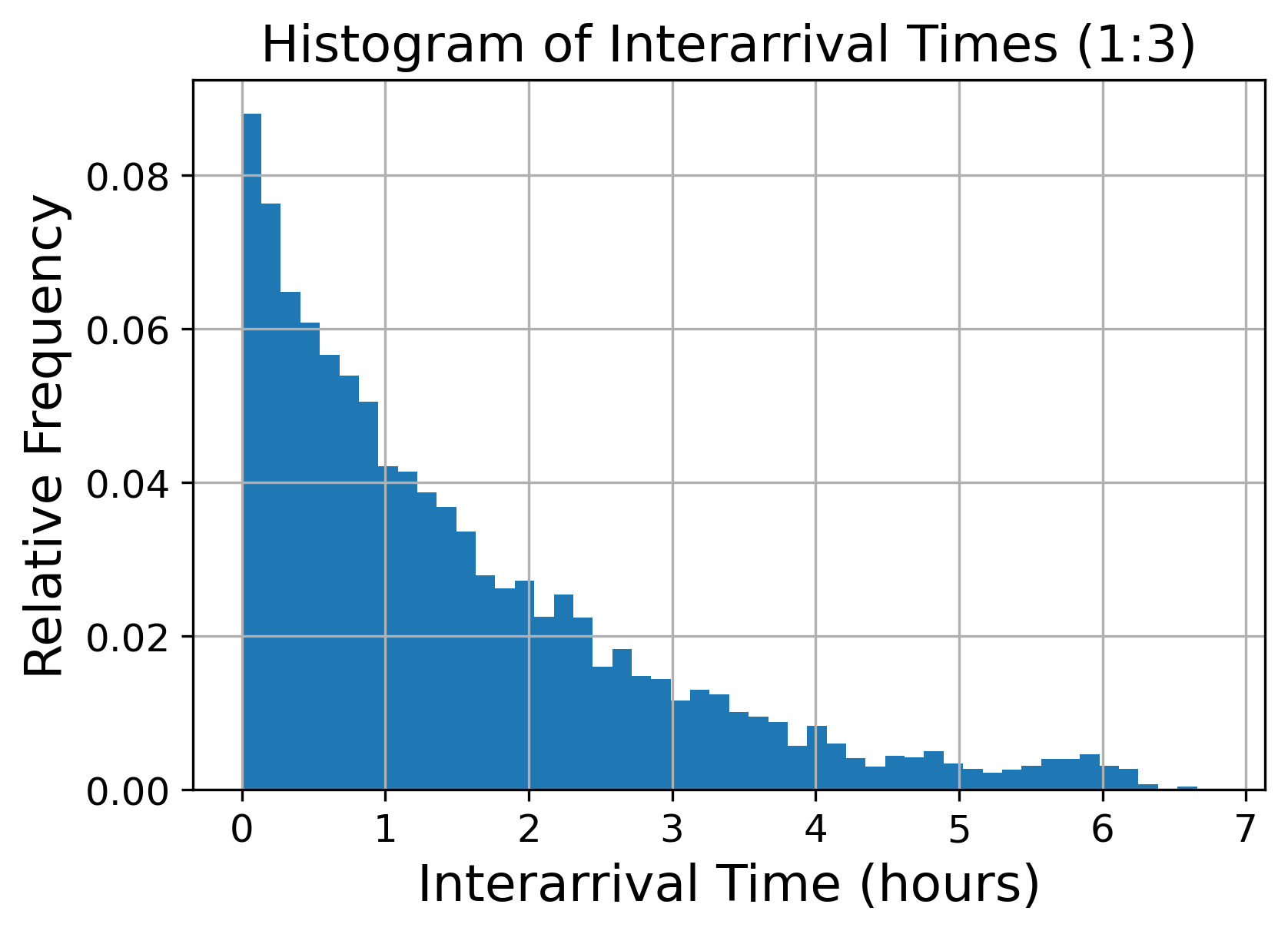}
    \end{subfigure}
        \hfill
    \begin{subfigure}{0.33\columnwidth}
        \centering
        \includegraphics[width=.95\linewidth]{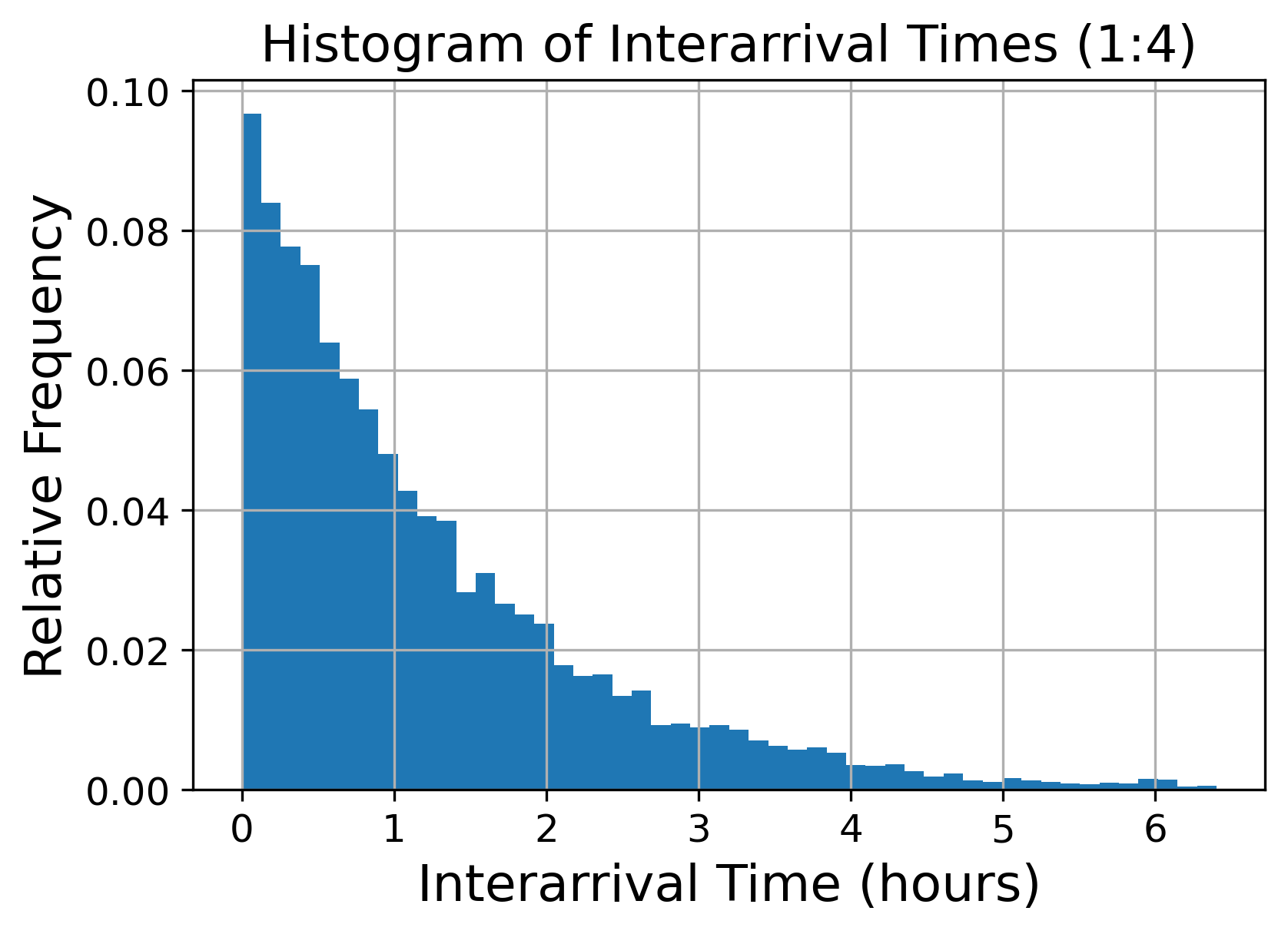}
    \end{subfigure}

    \caption{Histograms of interarrival times across different causal-to-noise ratios.}
    \label{fig:times}
\end{figure}

The distributions shown in Figure~\ref{fig:times} illustrate how the temporal structure of the event stream changes as the proportion of noise events increases. Without noise events, interarrival times remain concentrated around the mean causal delay of 6 hours, producing an approximately Gaussian-shaped distribution consistent with the synthetic generation process. 

As additional noise events from the independent Poisson process are introduced, the distribution progressively shifts toward shorter and more irregular temporal gaps, with increasing mass near small interarrival times. This transition disrupts the quasi-regular temporal structure of the causal stream, making temporal alignment substantially more difficult for standard lag-based causal discovery methods and motivating the need for time-aware approaches.

\subsection{Time Series of Features}
\label{sec:appendix_features}
\begin{figure}[H]
    \centering
    \begin{subfigure}{0.49\columnwidth}
        \centering
        \includegraphics[width=.9\linewidth]{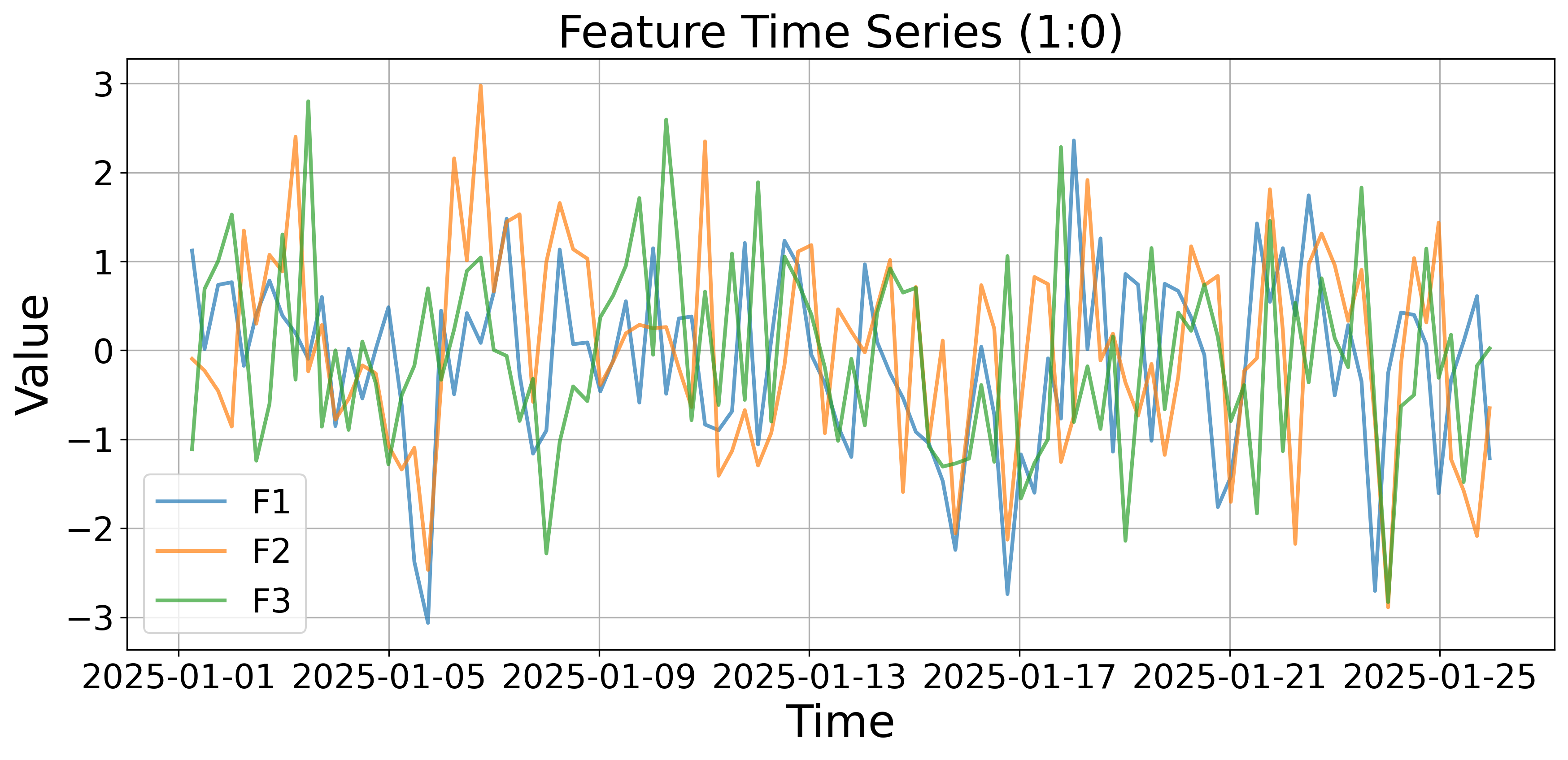}
    \end{subfigure}
    \hfill
    \begin{subfigure}{0.49\columnwidth}
        \centering
        \includegraphics[width=.9\linewidth]{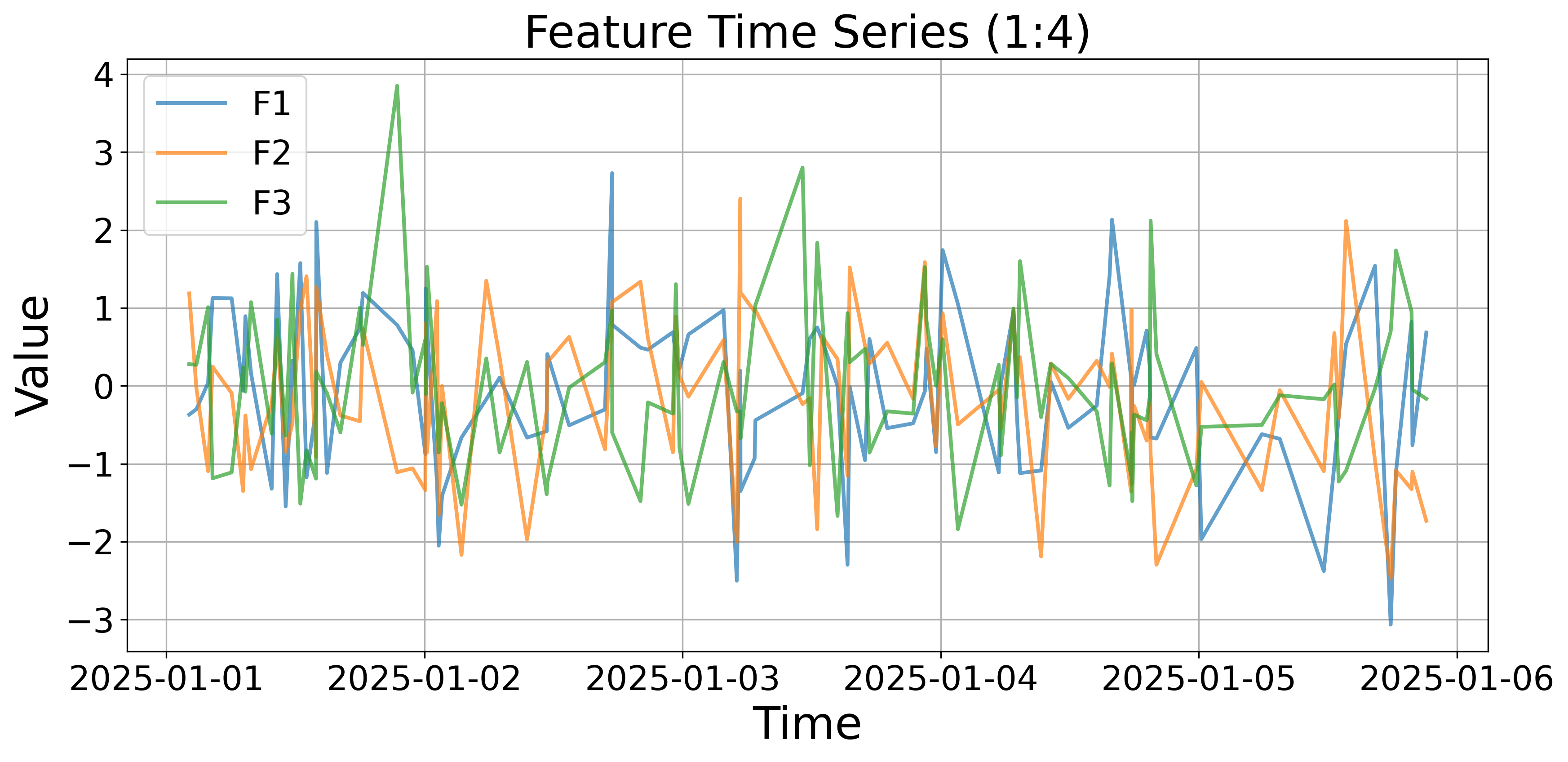}
    \end{subfigure}
    \centering
    \begin{subfigure}{0.49\columnwidth}
        \centering
        \includegraphics[width=.9\linewidth]{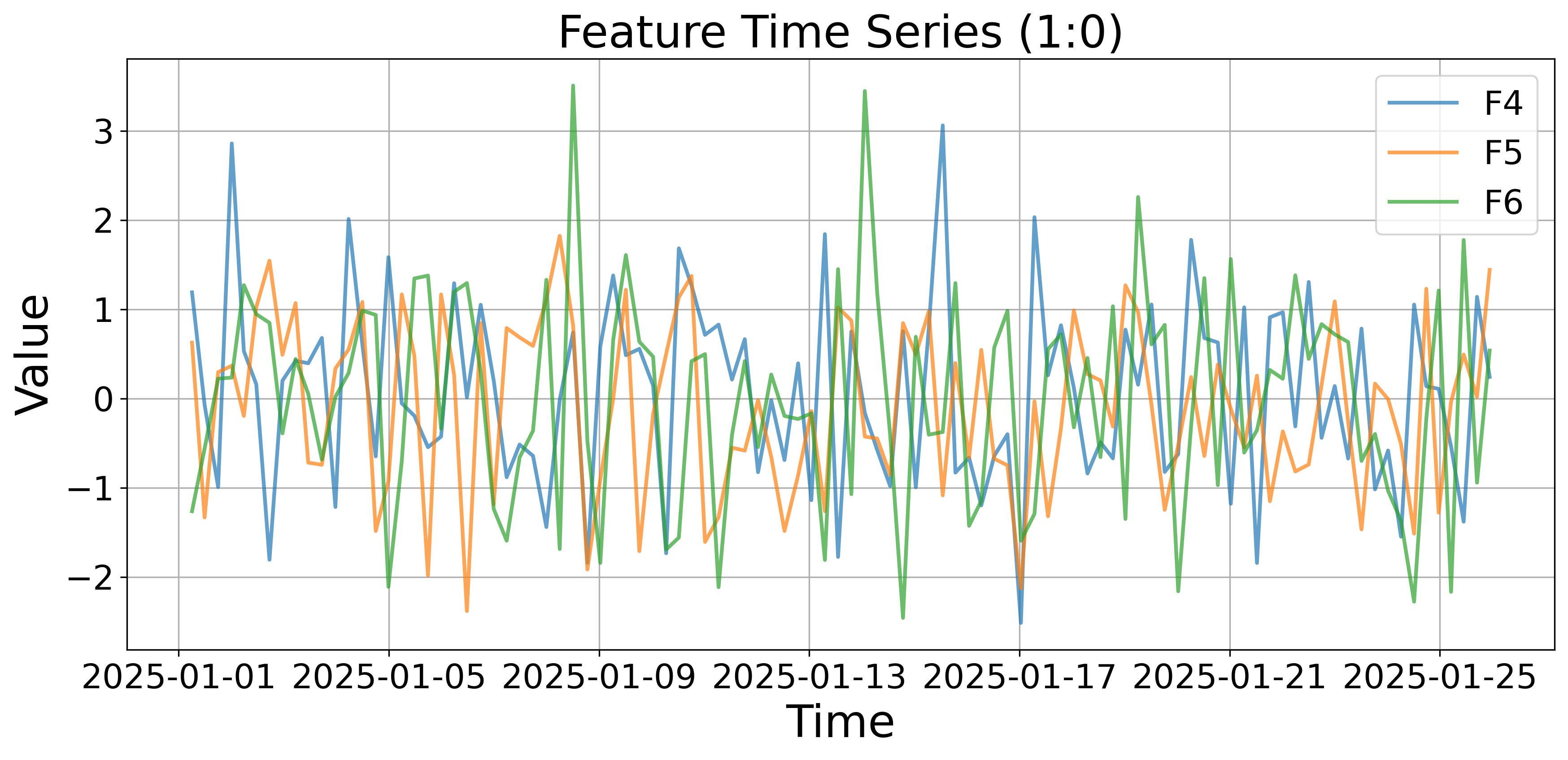}
    \end{subfigure}
    \hfill
    \begin{subfigure}{0.49\columnwidth}
        \centering
        \includegraphics[width=.9\linewidth]{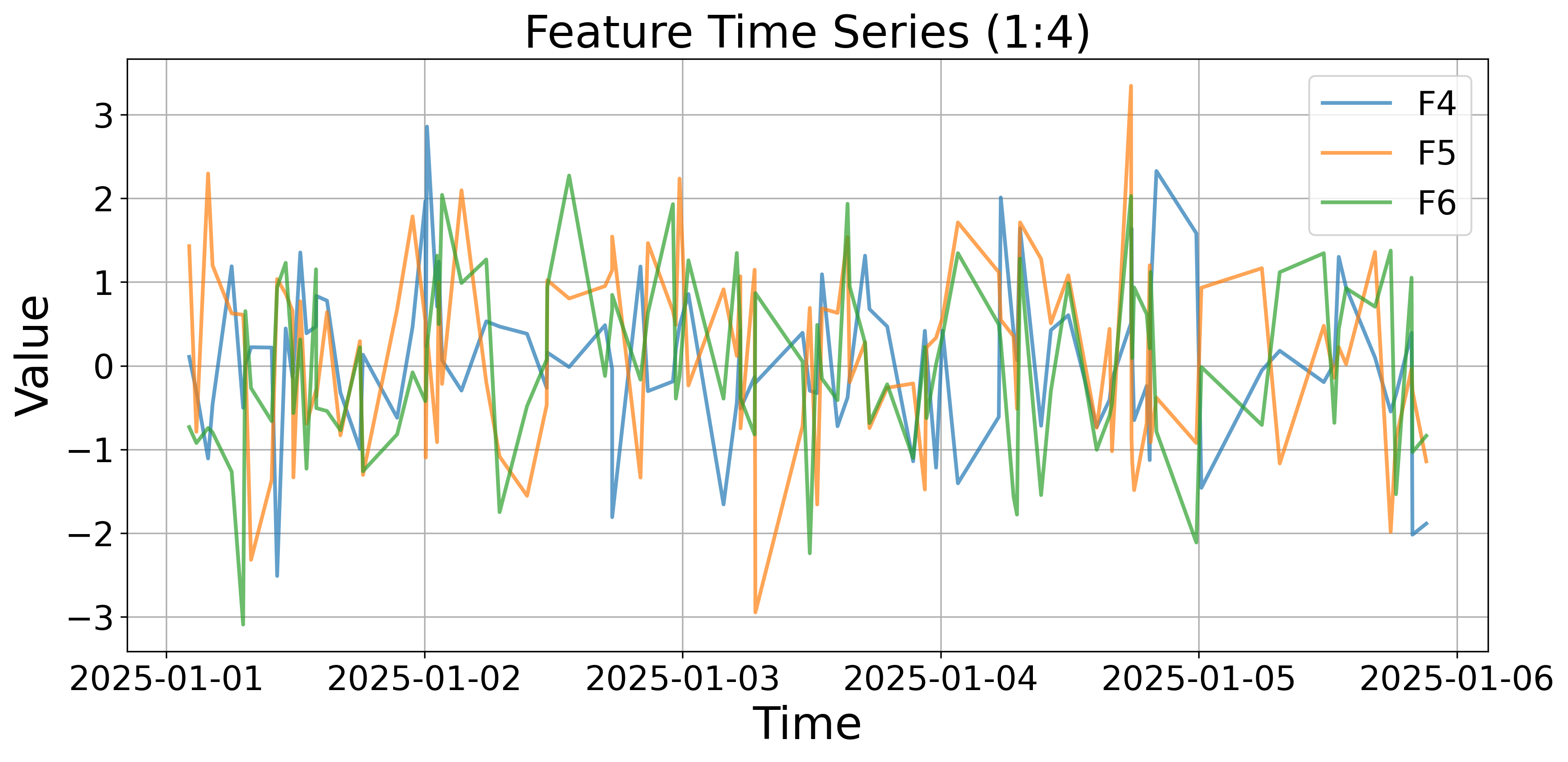}
    \end{subfigure}
       \centering
    \begin{subfigure}{0.49\columnwidth}
        \centering
        \includegraphics[width=.9\linewidth]{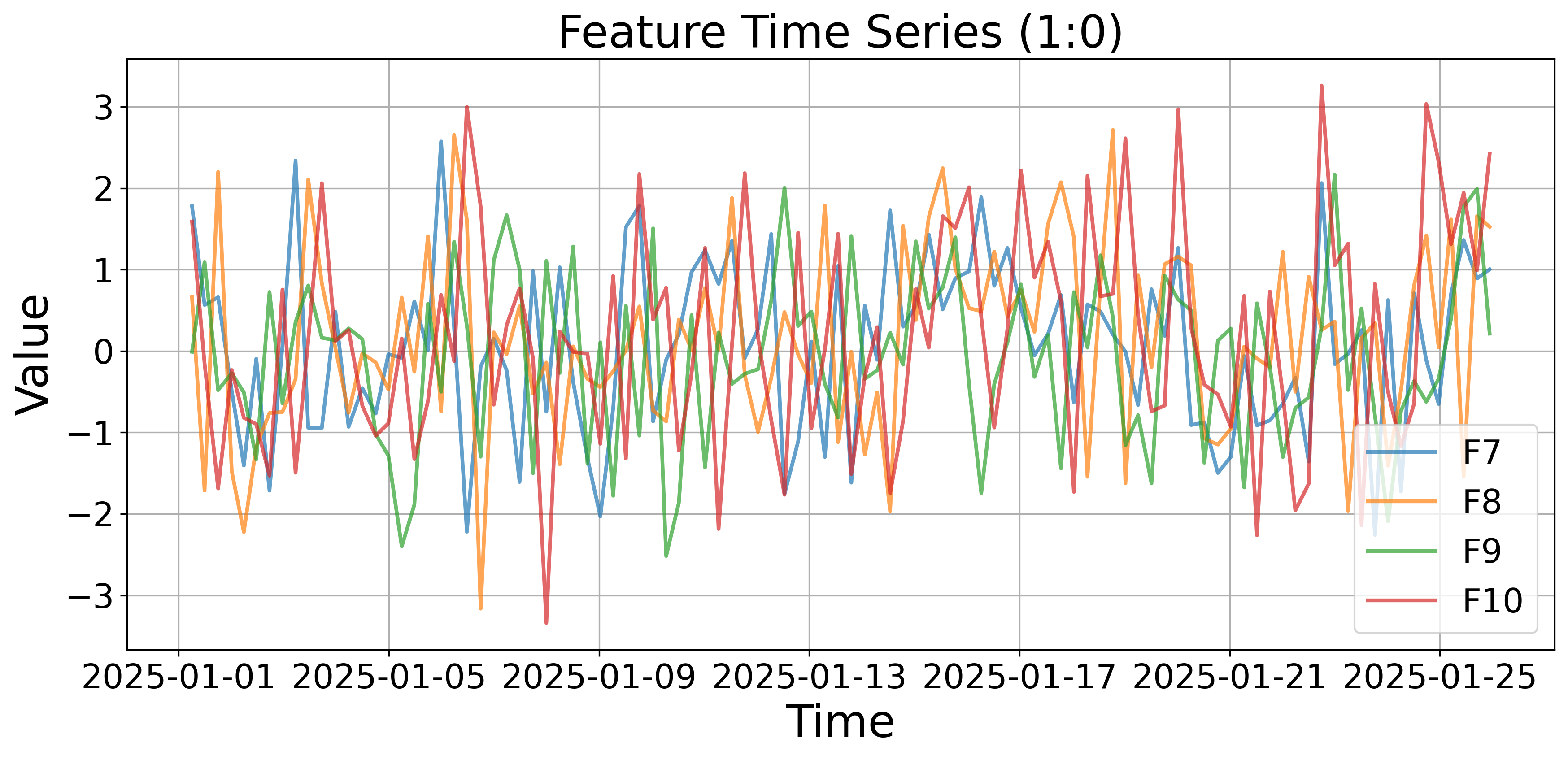}
    \end{subfigure}
    \hfill
    \begin{subfigure}{0.49\columnwidth}
        \centering
        \includegraphics[width=.9\linewidth]{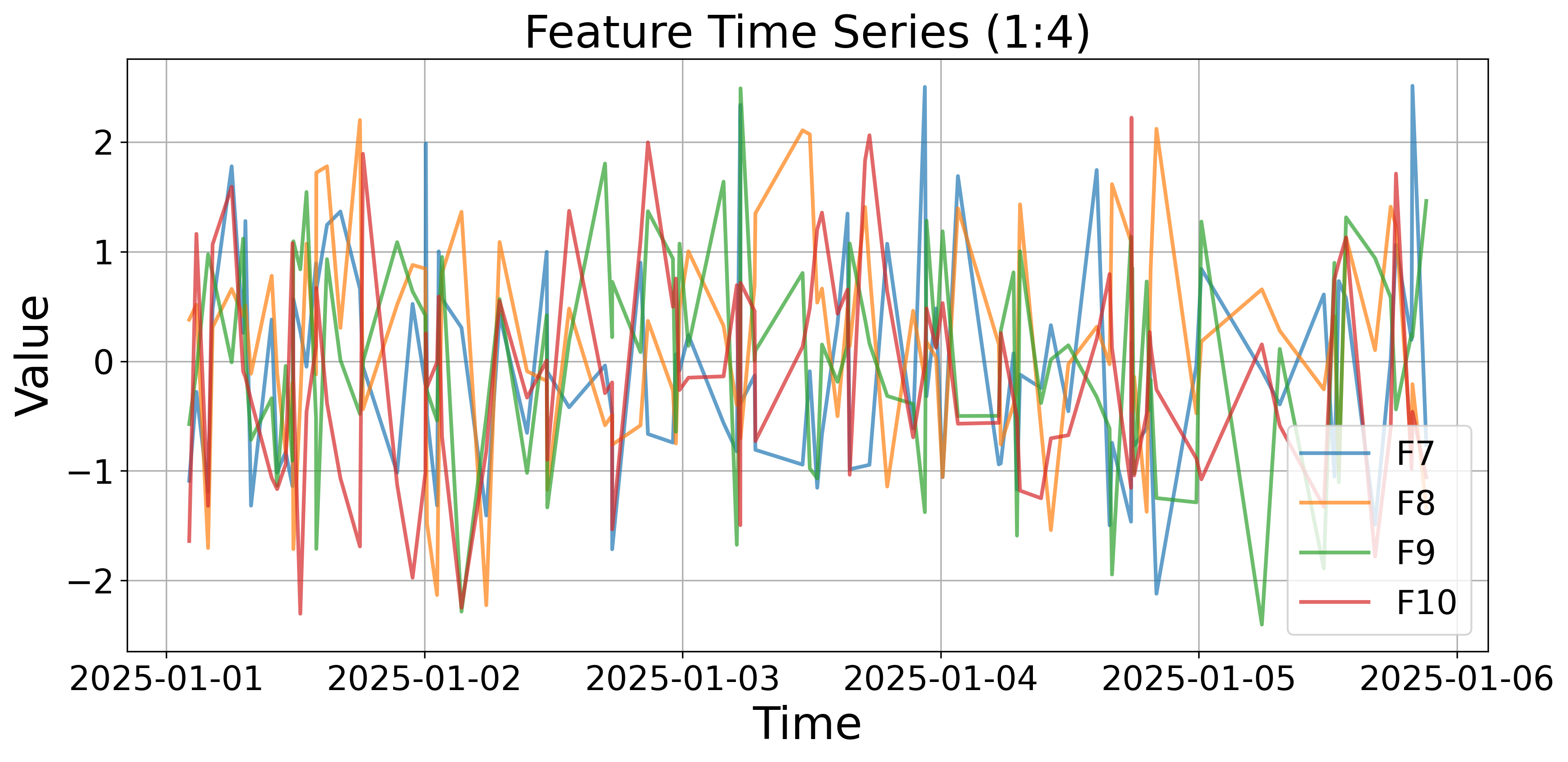}
    \end{subfigure}
           \centering
    \begin{subfigure}{0.49\columnwidth}
        \centering
        \includegraphics[width=.9\linewidth]{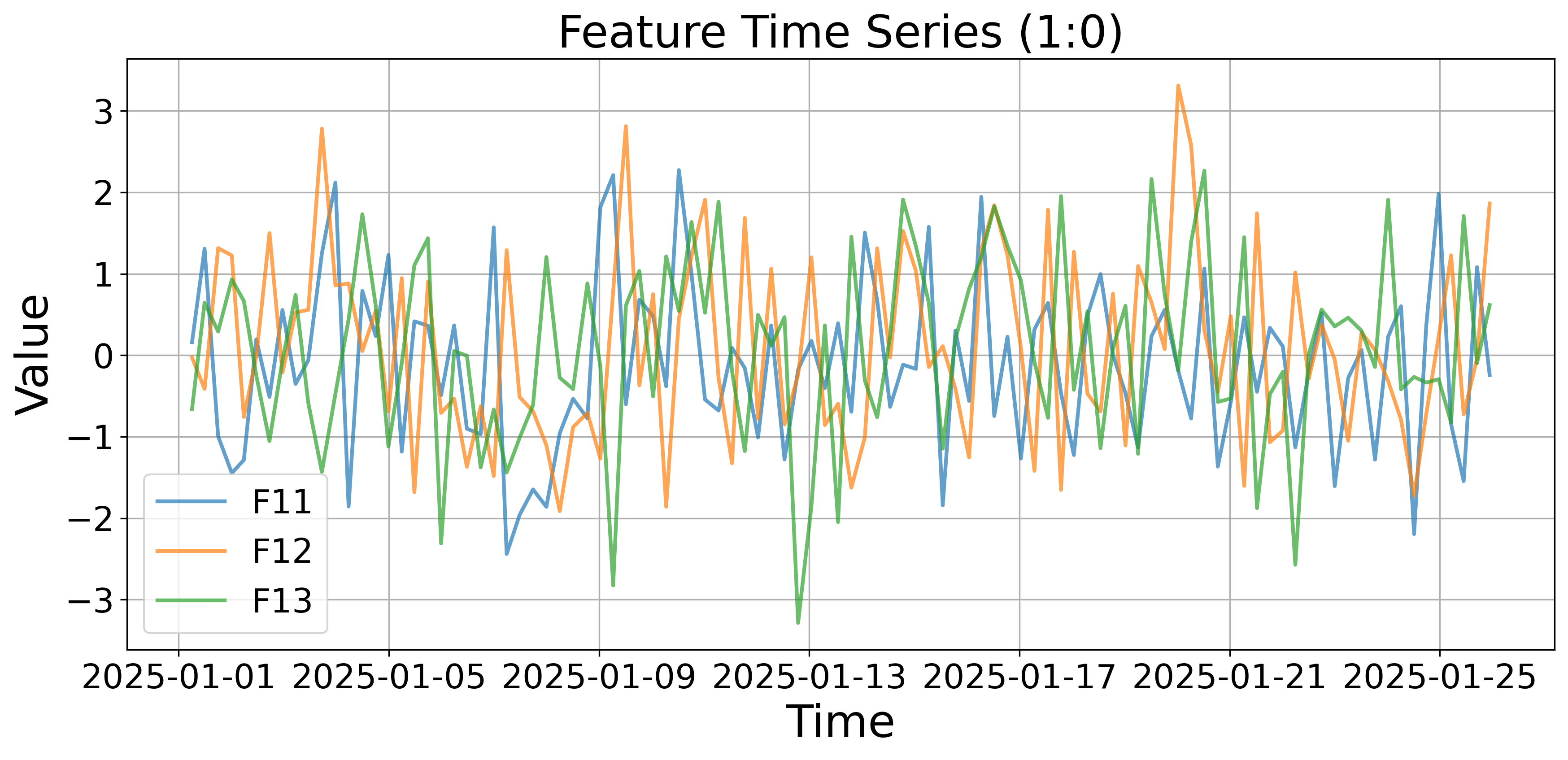}
    \end{subfigure}
    \hfill
    \begin{subfigure}{0.49\columnwidth}
        \centering
        \includegraphics[width=.9\linewidth]{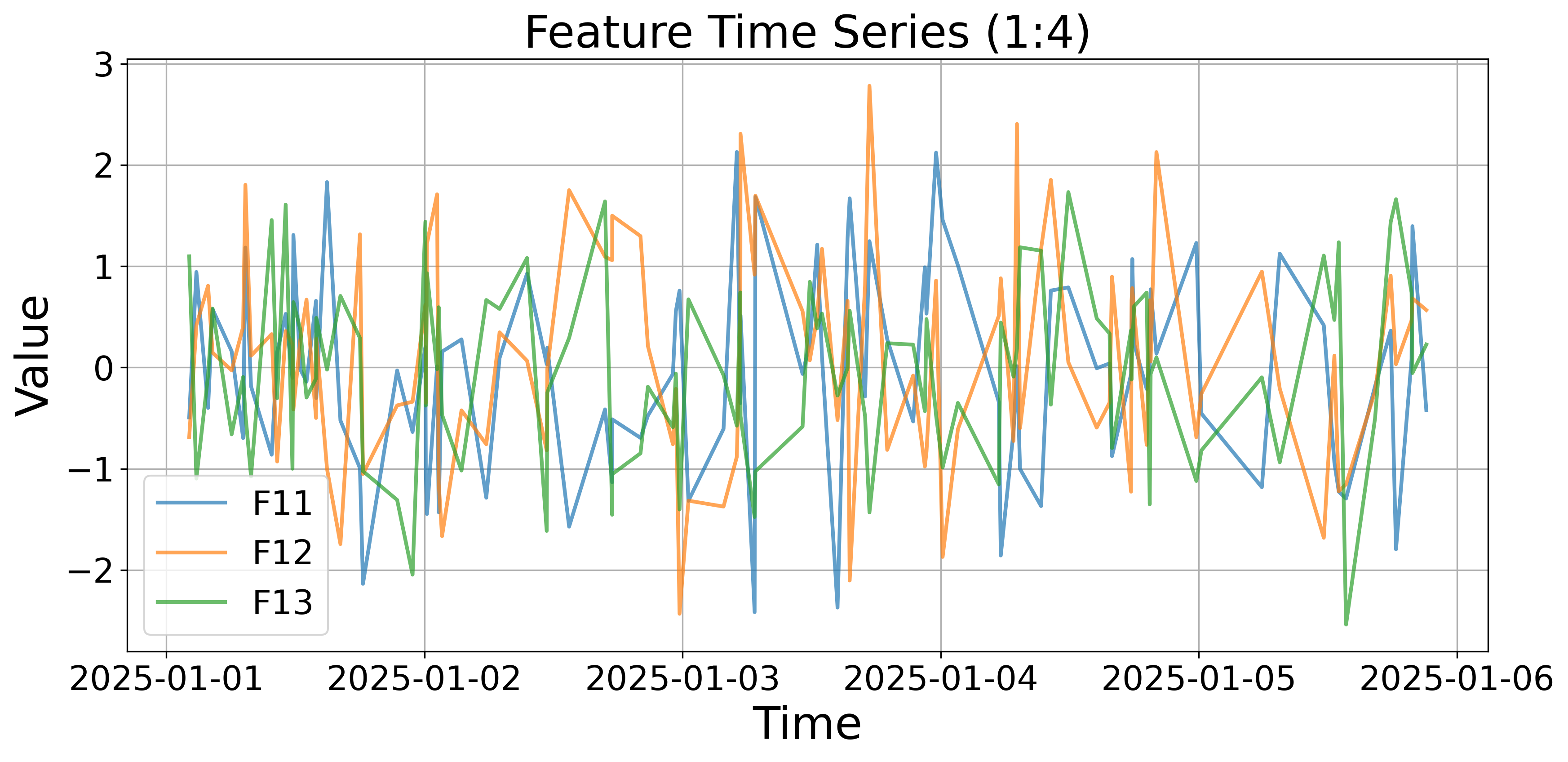}
    \end{subfigure}
           \centering
    \begin{subfigure}{0.49\columnwidth}
        \centering
        \includegraphics[width=.9\linewidth]{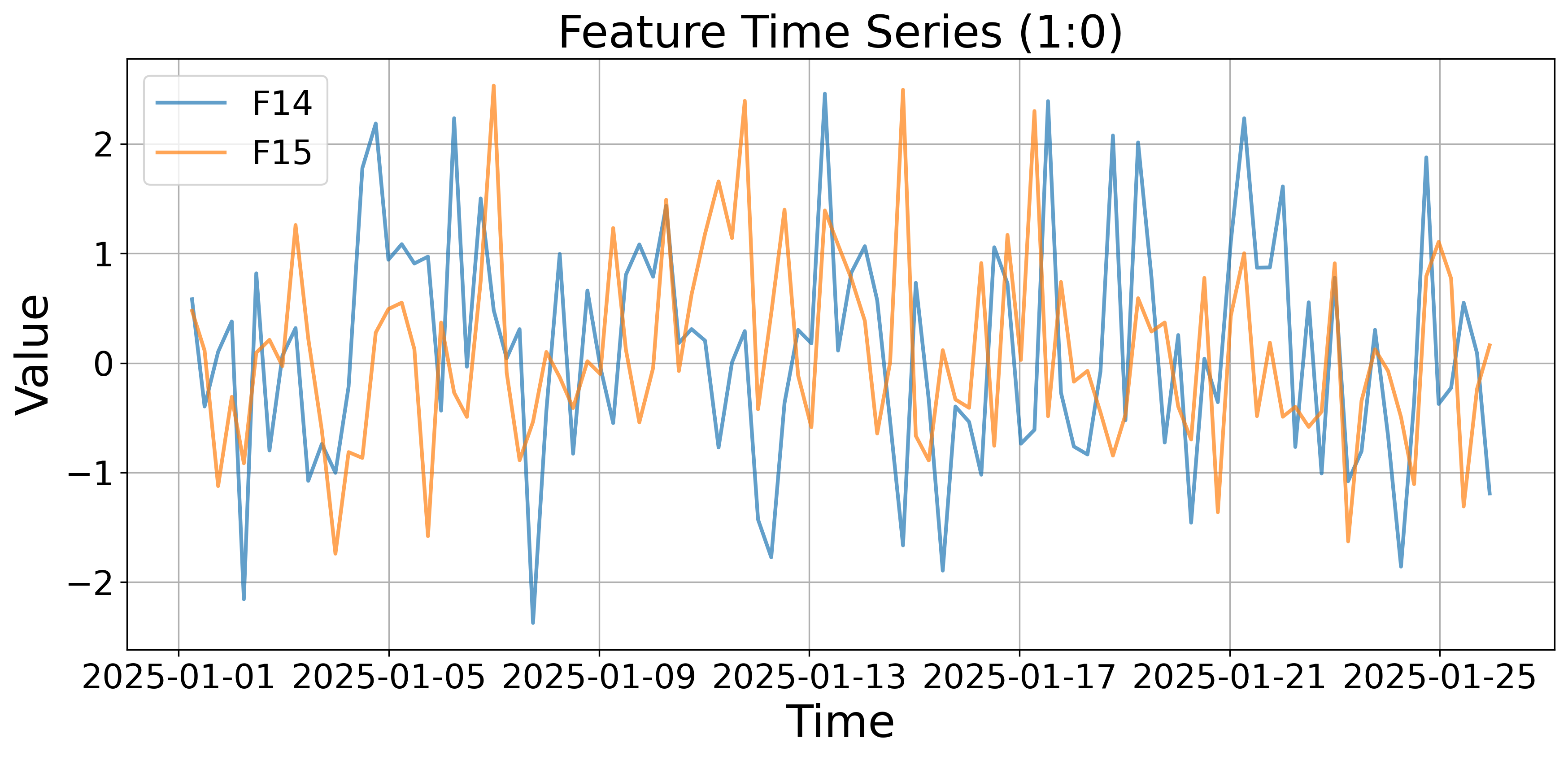}
    \end{subfigure}
    \hfill
    \begin{subfigure}{0.49\columnwidth}
        \centering
        \includegraphics[width=.9\linewidth]{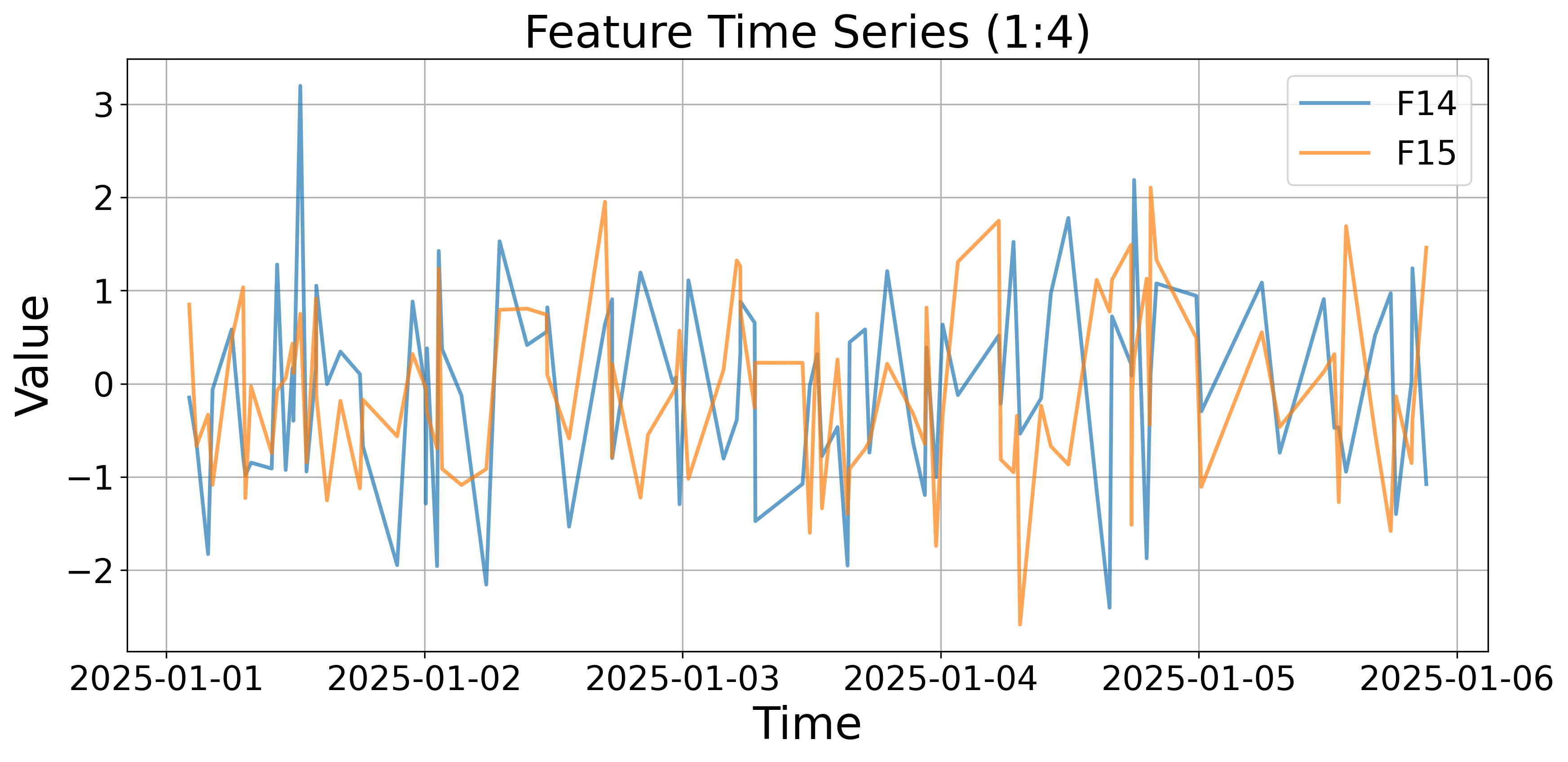}
    \end{subfigure}
       \caption{Time series of features, grouped according to their underlying causal structures, for the first 100 events, under causal-to-noise ratios of 1:0 and 1:4.}
    \label{fig:timeseries}
\end{figure}

The time series shown in Figure~\ref{fig:timeseries} illustrate variables generated from different temporal causal structures: a fork (\(F_1\)–\(F_3\)), a collider (\(F_4\)–\(F_6\)), a diamond (\(F_7\)–\(F_{10}\)), and a causal chain (\(F_{11}\)–\(F_{13}\)), while \(F_{14}\) and \(F_{15}\) correspond to independent noise variables. At low noise levels, variables within the same causal structure exhibit coordinated temporal dynamics. As the number of noise events increases, these dependencies become progressively less distinguishable, illustrating the challenges posed by irregular and noisy event streams.

\section{Window Configurations}
\label{sec:appendix_windows}
We illustrate the temporal window configurations used throughout the experiments presented in Section~\ref{sec:results}. 
The figures provide a visual interpretation of how lag representations are constructed under both the rectangular-window and Gaussian-window variants of the proposed framework.

\begin{figure}[H]
    \centering
    \begin{subfigure}{0.49\columnwidth}
        \centering
        \includegraphics[width=\linewidth]{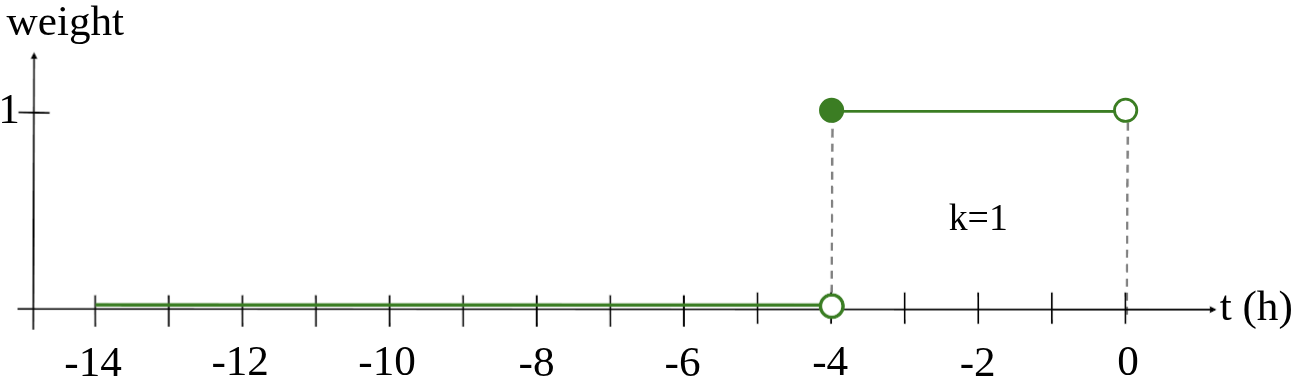}
    \end{subfigure}
    \hfill
    \begin{subfigure}{0.49\columnwidth}
        \centering
        \includegraphics[width=\linewidth]{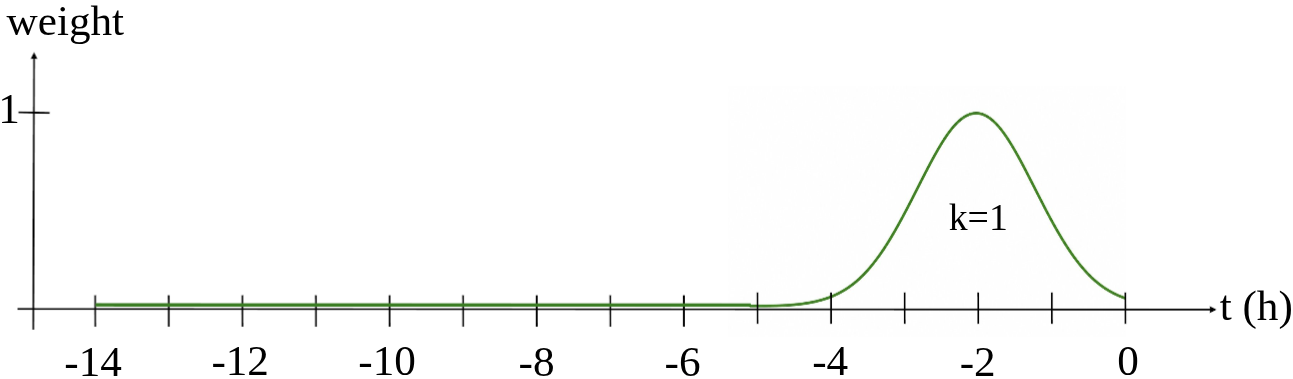}
    \end{subfigure}
    
    \vspace{3mm}
    
    \begin{subfigure}{0.49\columnwidth}
        \centering
        \includegraphics[width=\linewidth]{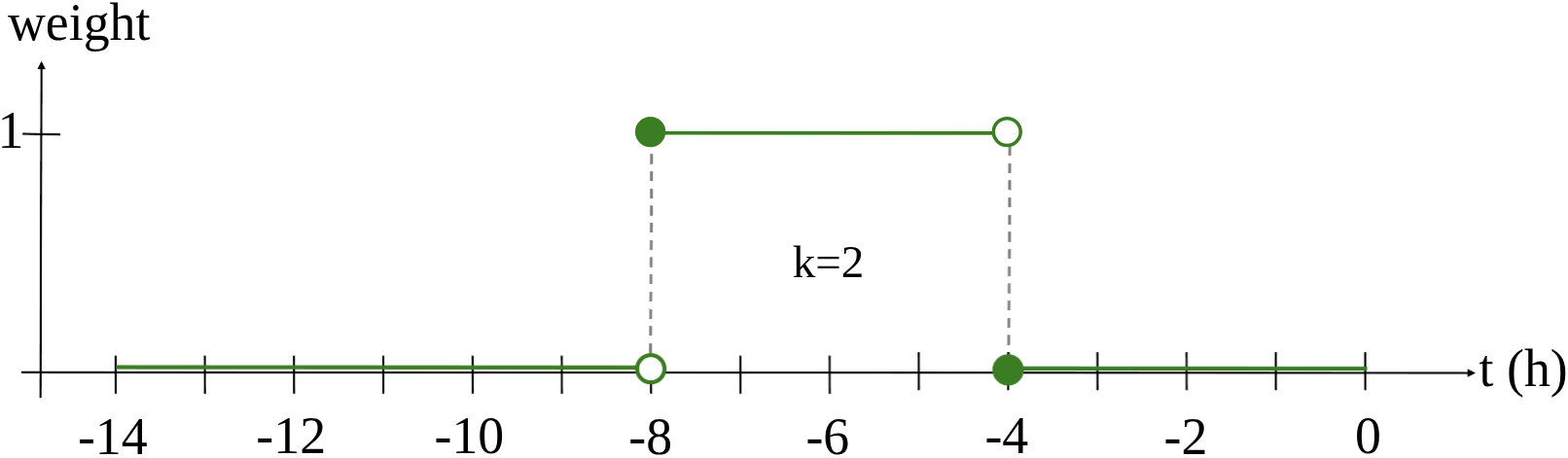}
    \end{subfigure}
    \hfill
    \begin{subfigure}{0.49\columnwidth}
        \centering
        \includegraphics[width=\linewidth]{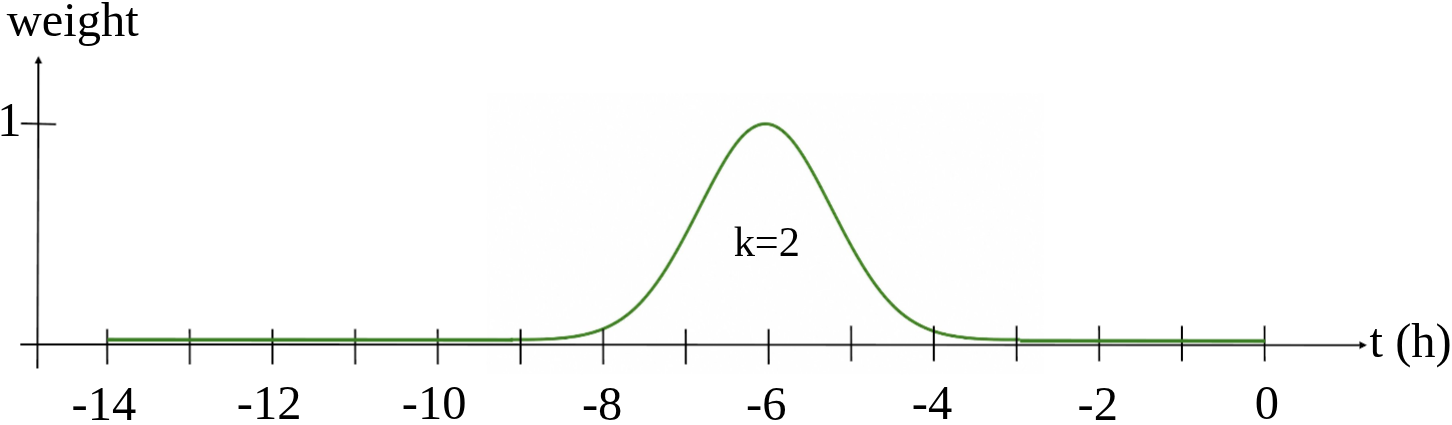}
    \end{subfigure}
    
    \vspace{3mm}
    
        \centering
    \begin{subfigure}{0.49\columnwidth}
        \centering
        \includegraphics[width=\linewidth]{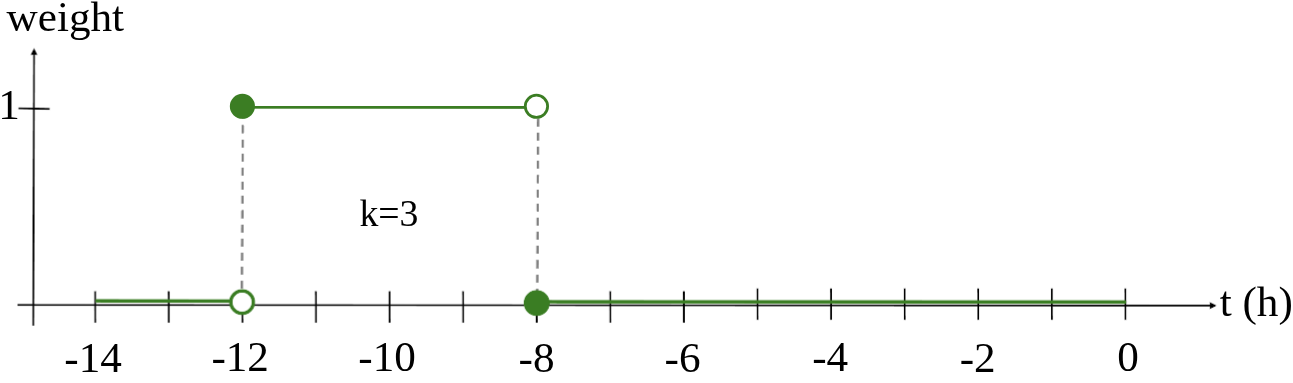}
    \end{subfigure}
    \hfill
    \begin{subfigure}{0.49\columnwidth}
        \centering
        \includegraphics[width=\linewidth]{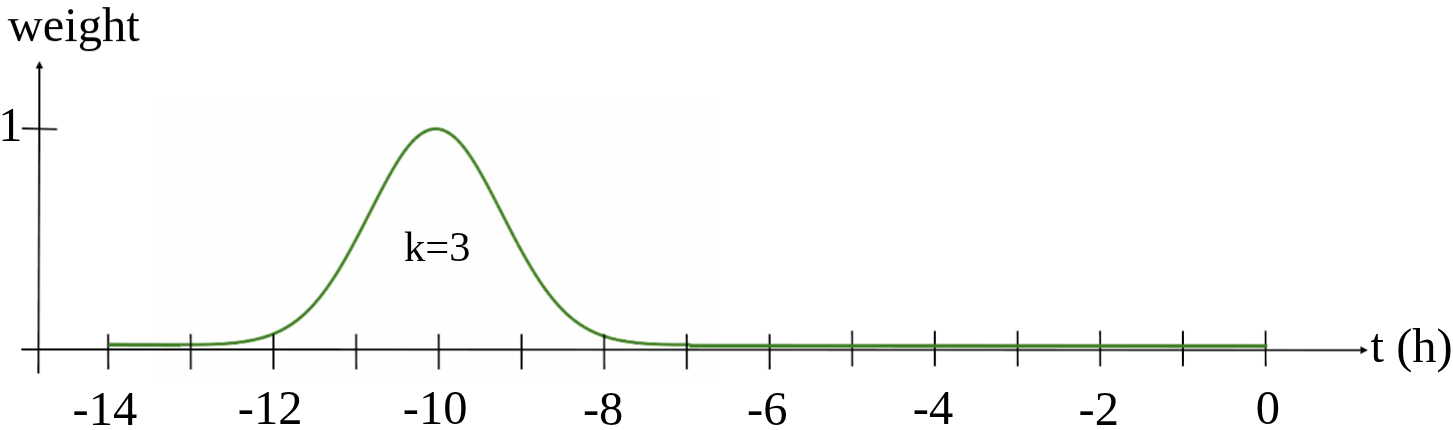}
    \end{subfigure}
     \caption{
    Comparison between rectangular-window (left column) and Gaussian-window (right column) temporal weights for the lag configurations of Experiment 1. 
    Rectangular windows assign equal weight to all observations inside the interval, whereas Gaussian windows weight observations according to their temporal distance from the target delay.
    }
    \label{fig:window_shapes}
\end{figure}

In the rectangular-window formulation, each lag corresponds to a finite temporal interval in which all observations receive equal weight. 
The left column of Figure~\ref{fig:window_shapes} illustrates the assignment regions for the lag configurations of Experiment 1. 
In contrast, the Gaussian-window formulation replaces the binary inclusion rule with a continuous weighting function centred around the target temporal delay. The right column of Figure~\ref{fig:window_shapes} shows the corresponding Gaussian weighting profiles for the same experiment, where observations closer to the centre of the window contribute more strongly than distant observations.

\begin{figure}[H]
    \centering
    \includegraphics[width=0.9\textwidth]{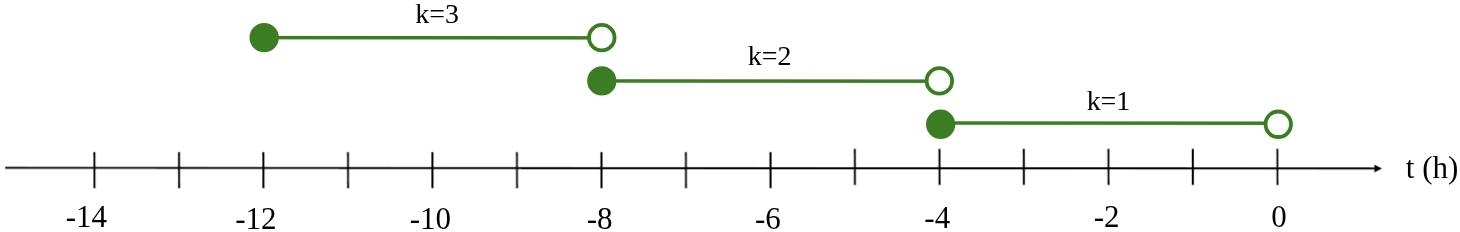}
    \caption{Illustration of the window configurations utilized in Experiment 1.}
    \label{fig:windows_1}
\end{figure}

\begin{figure}[H]
    \centering
    \includegraphics[width=0.9\textwidth]{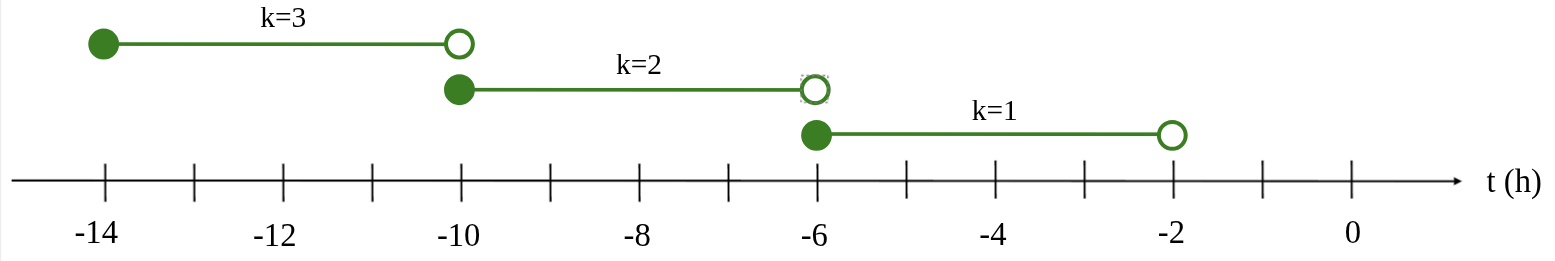}
    \caption{Illustration of the window configurations utilized in Experiment 2.}
    \label{fig:windows_2}
\end{figure}

\begin{figure}[H]
    \centering
    \includegraphics[width=0.9\textwidth]{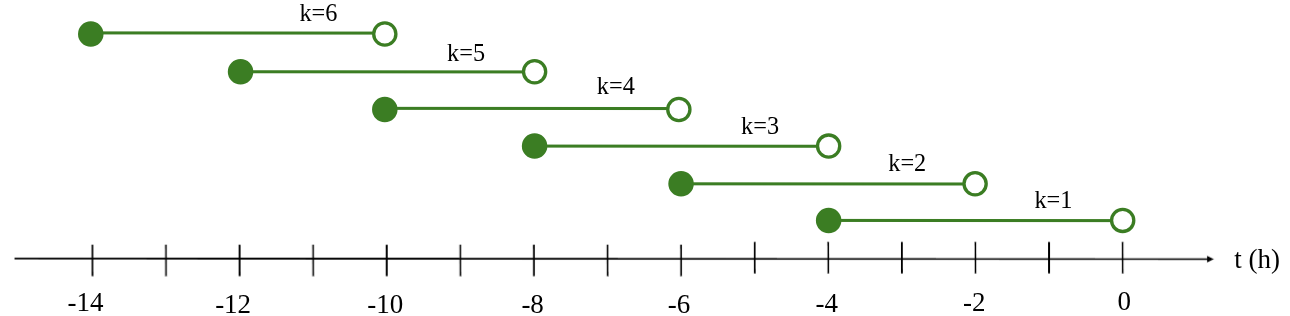}
    \caption{Illustration of the window configurations utilized in Experiment 3.}
    \label{fig:windows_3}
\end{figure}

\begin{figure}[H]
    \centering
    \includegraphics[width=0.9\textwidth]{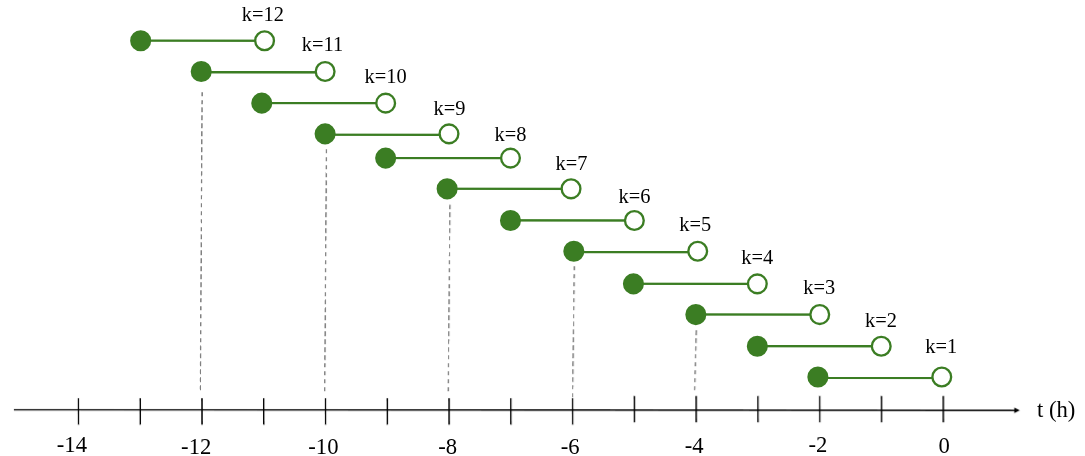}
    \caption{Illustration of the window configurations utilized in Experiment 4.}
    \label{fig:windows_4}
\end{figure}

Figures~\ref{fig:windows_1}--\ref{fig:windows_4} illustrate the specific temporal configurations used in Experiments 1-4.

\section{Additional simulation results}
\label{sec:appendixA}
\subsection{Recall and Precision of all edges}
\begin{figure}
    \centering
    \begin{subfigure}{0.24\columnwidth}
        \centering
        \includegraphics[width=\linewidth]{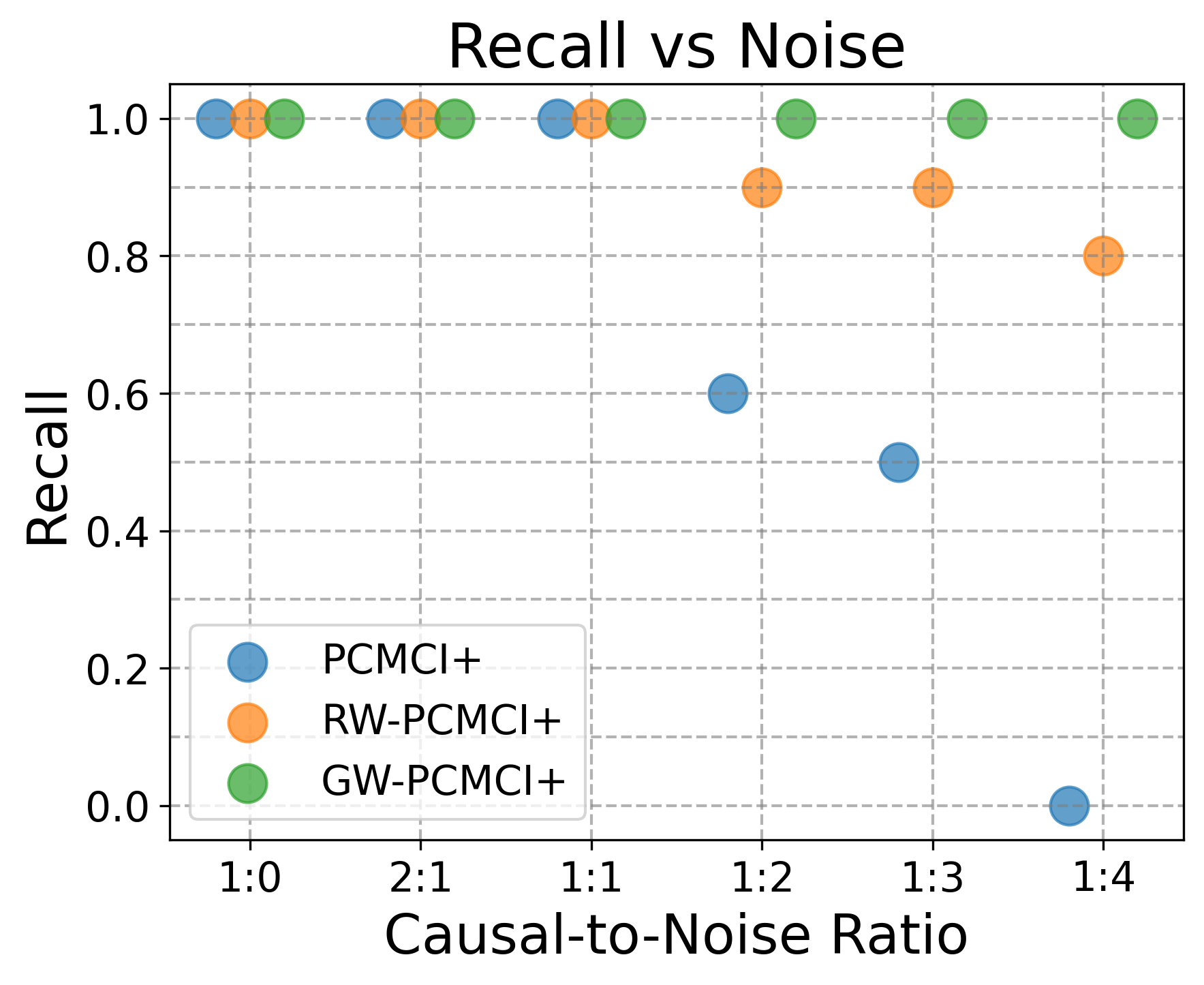}
    \end{subfigure}
    \hfill
    \begin{subfigure}{0.24\columnwidth}
        \centering
        \includegraphics[width=\linewidth]{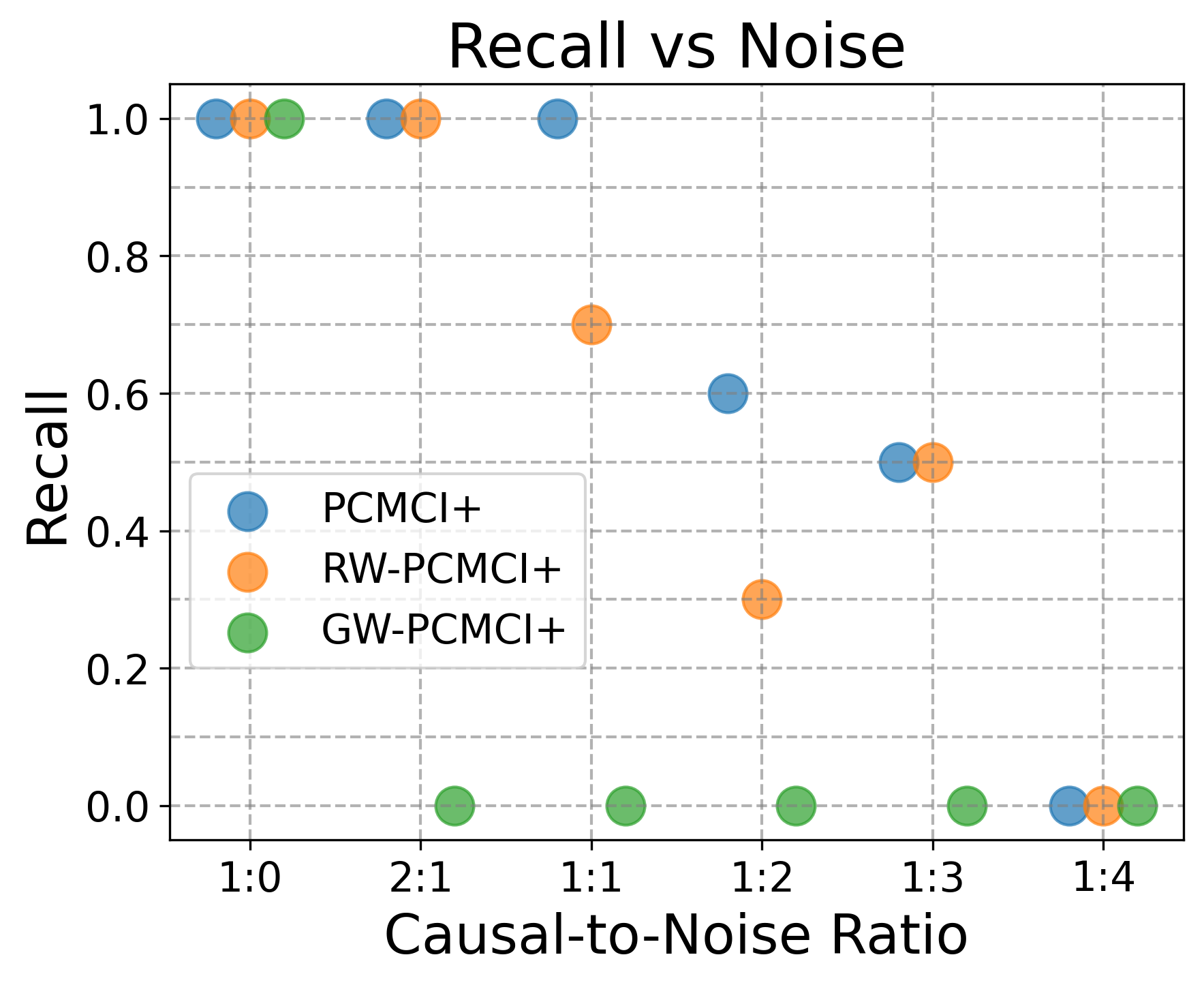}
    \end{subfigure}
    \centering
    \begin{subfigure}{0.24\columnwidth}
        \centering
        \includegraphics[width=\linewidth]{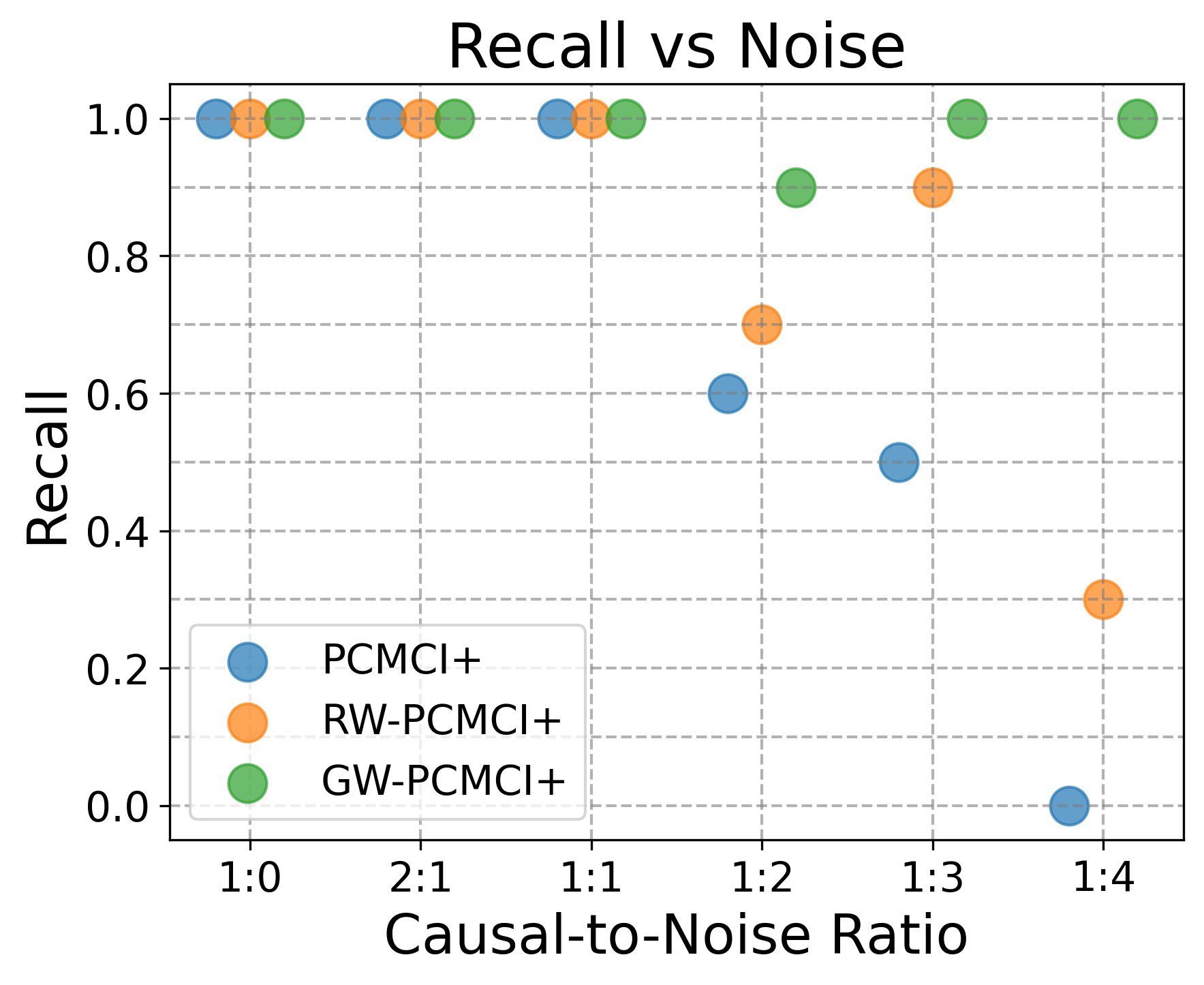}
    \end{subfigure}
    \hfill
    \begin{subfigure}{0.24\columnwidth}
        \centering
        \includegraphics[width=\linewidth]{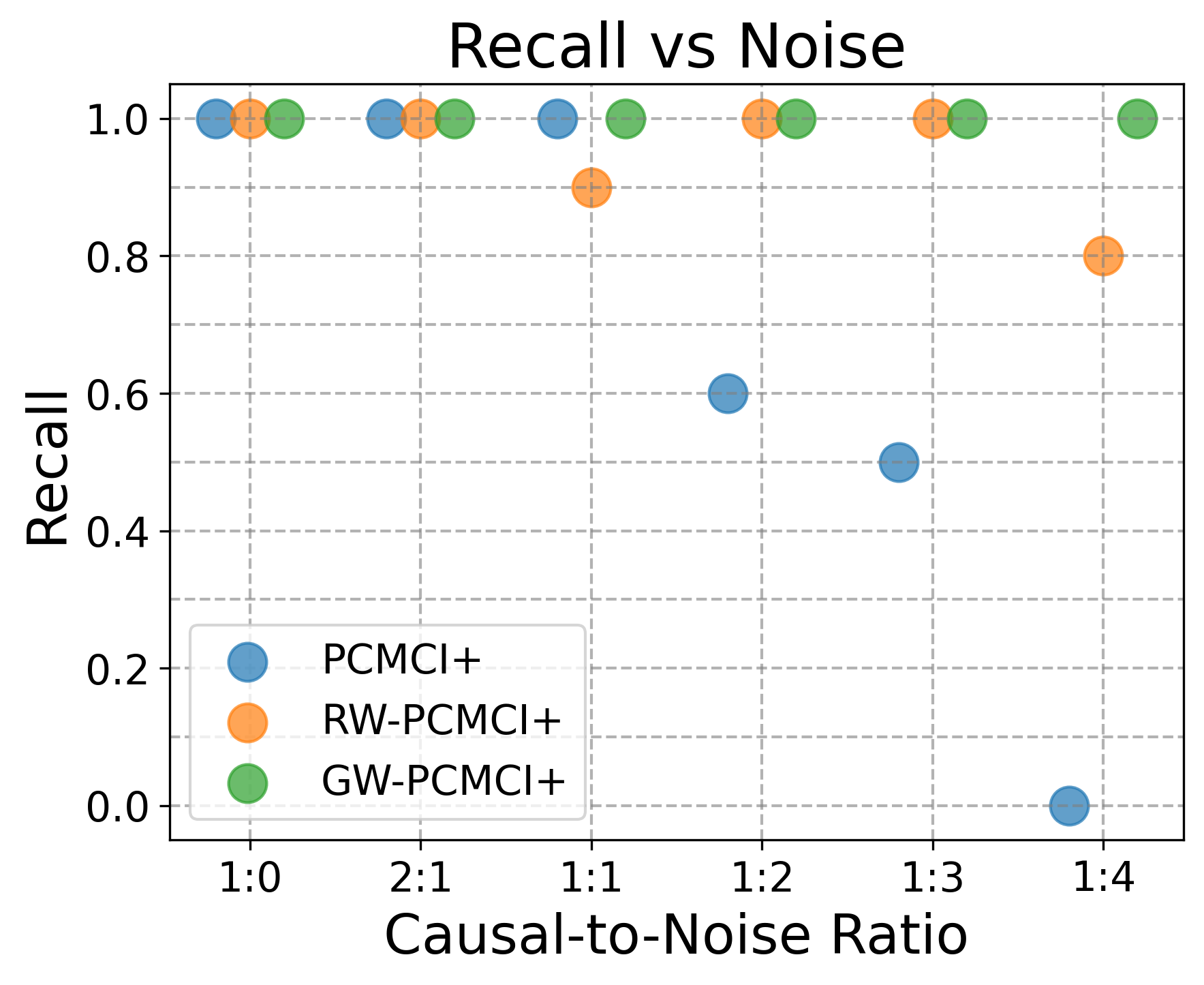}
    \end{subfigure}
       \centering
    \begin{subfigure}{0.24\columnwidth}
        \centering
        \includegraphics[width=\linewidth]{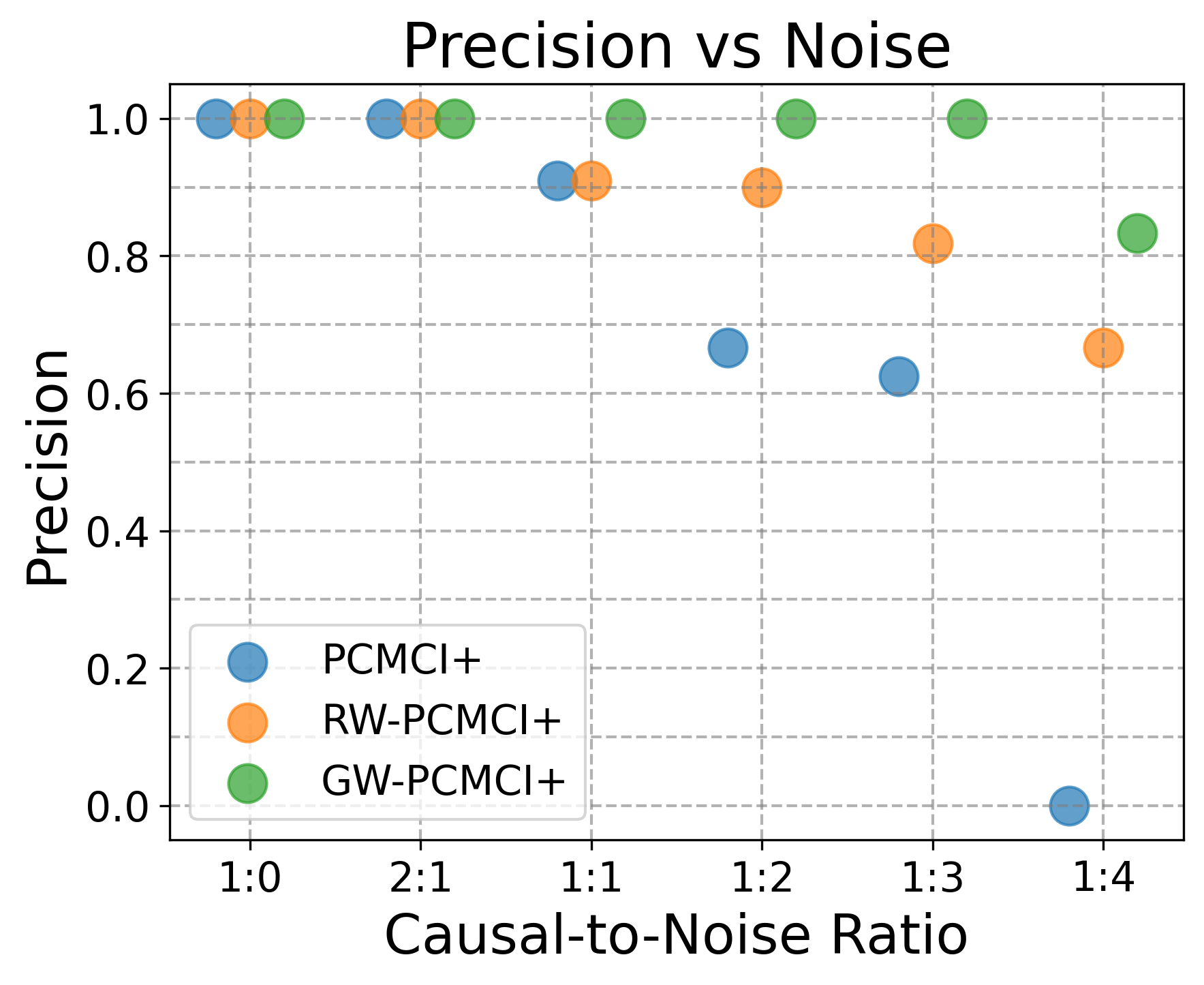}
    \end{subfigure}
    \hfill
    \begin{subfigure}{0.24\columnwidth}
        \centering
        \includegraphics[width=\linewidth]{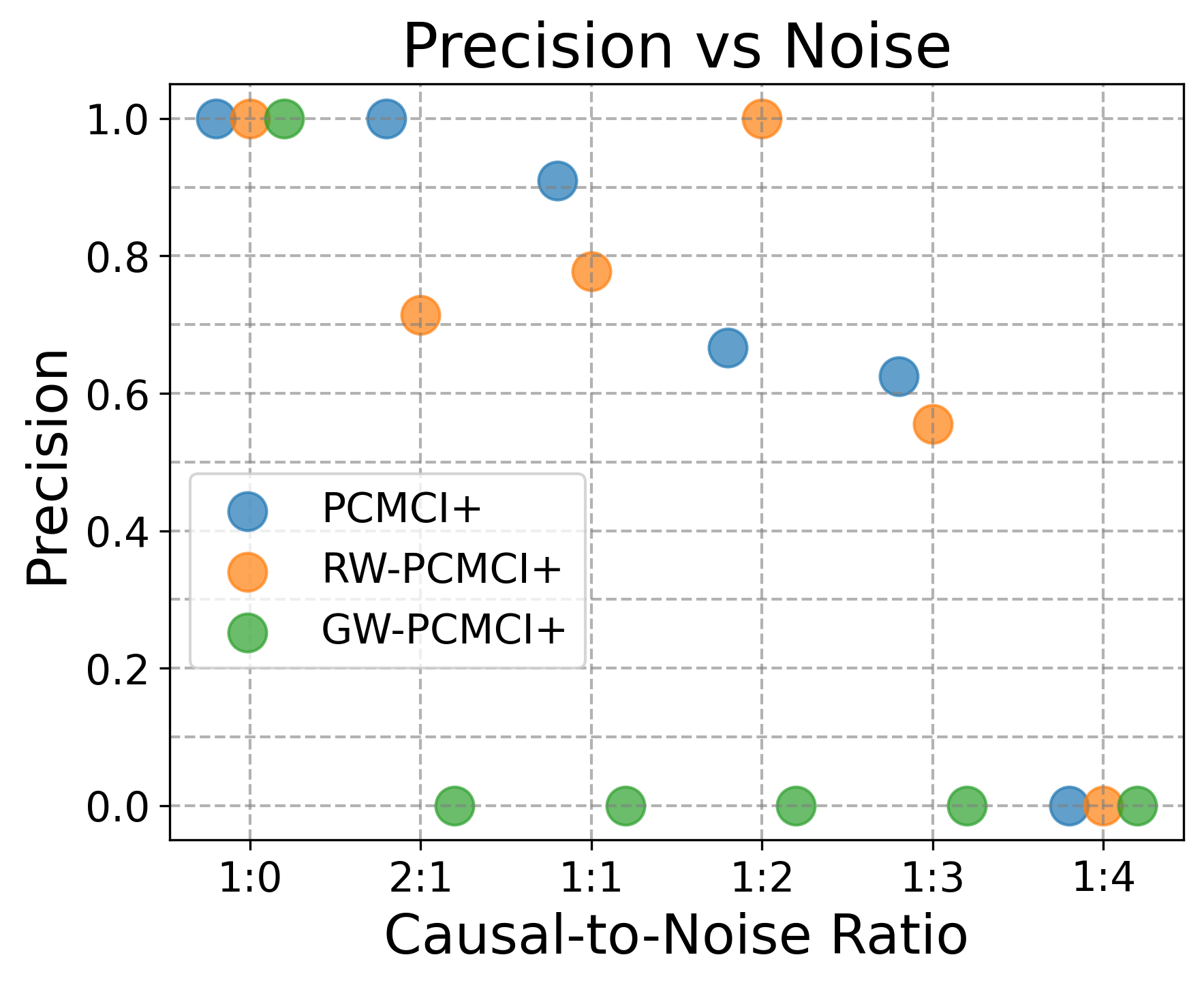}
    \end{subfigure}
           \centering
    \begin{subfigure}{0.24\columnwidth}
        \centering
        \includegraphics[width=\linewidth]{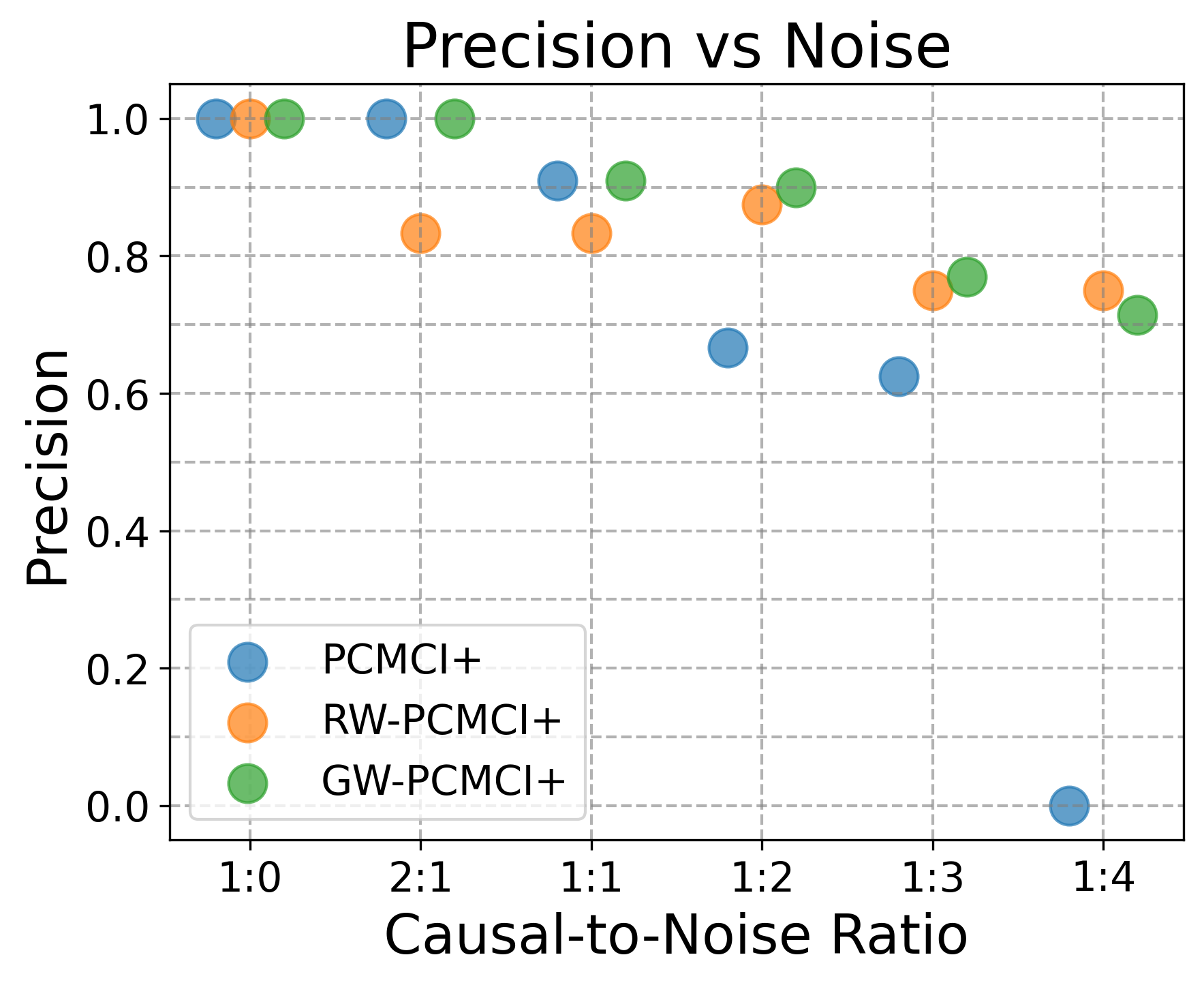}
    \end{subfigure}
    \hfill
    \begin{subfigure}{0.24\columnwidth}
        \centering
        \includegraphics[width=\linewidth]{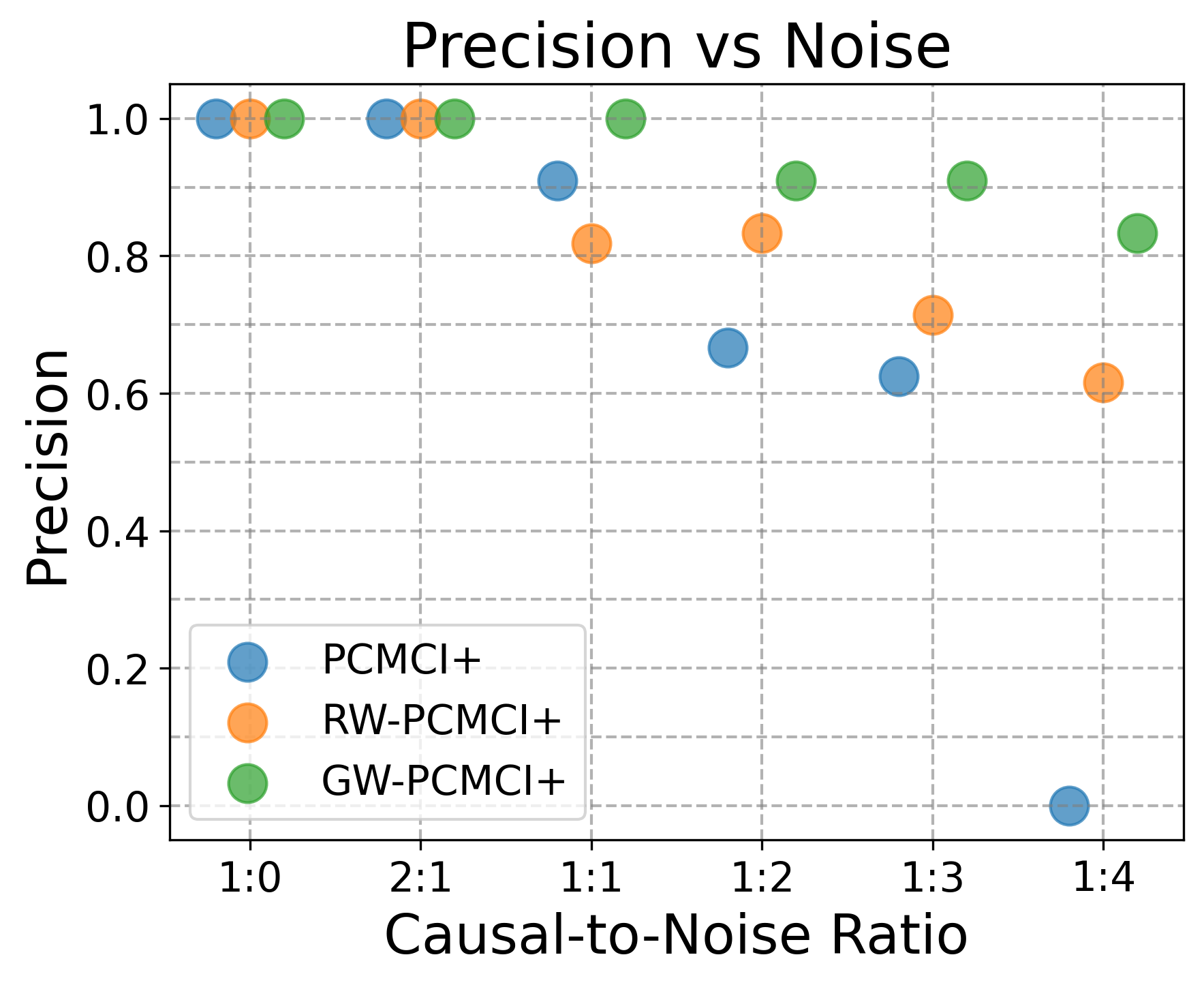}
    \end{subfigure}
         \caption{Recall and precision for Experiments 1-4 (first through fourth columns, respectively) across varying causal-to-noise ratios.}
    \label{fig:rec_prec}
\end{figure}

In addition to SHD, we further report the proposed methods using recall and precision metrics across all experimental settings. While SHD provides a global measure of graph recovery quality, recall and precision offer complementary insights into the types of errors made by each method. Recall measures the ability to recover the true causal edges present in the underlying graph, whereas precision measures the proportion of inferred edges that correspond to true causal relationships rather than false positives. Together, these metrics provide a more detailed characterization of method performance under increasing levels of noise.
As we can see in Figure~\ref{fig:rec_prec}, the Gaussian-window extension notably achieves maximum recall in Experiments 1, 3, and 4 at every causal-to-noise ratio, indicating that it consistently recovers all true edges.

\subsection{SHD, Recall and Precision of Lagged Edges}
\begin{figure}
\centering
    \begin{subfigure}{0.24\columnwidth}
        \centering
        \includegraphics[width=\linewidth]{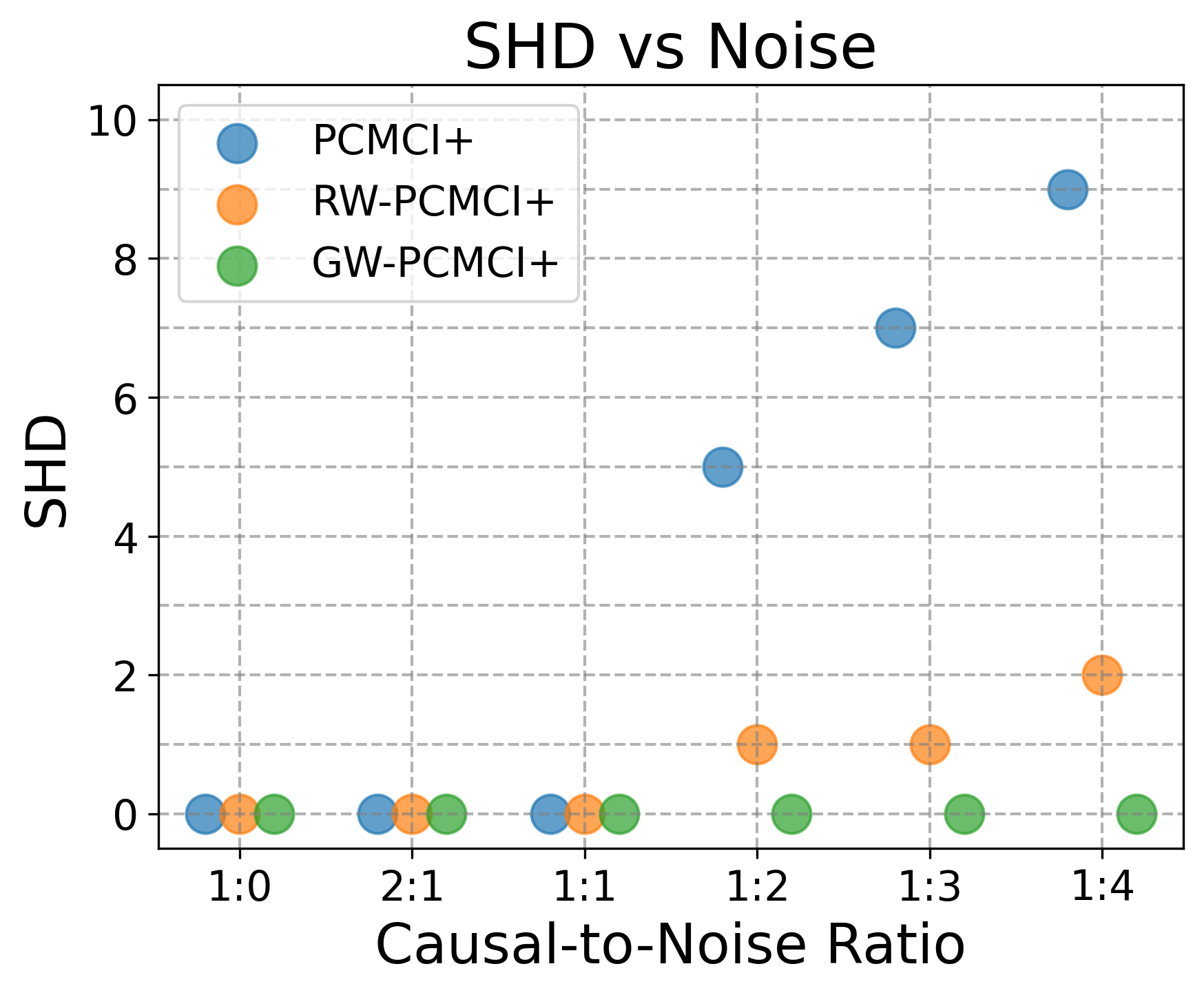}
    \end{subfigure}
    \hfill
    \begin{subfigure}{0.24\columnwidth}
        \centering
        \includegraphics[width=\linewidth]{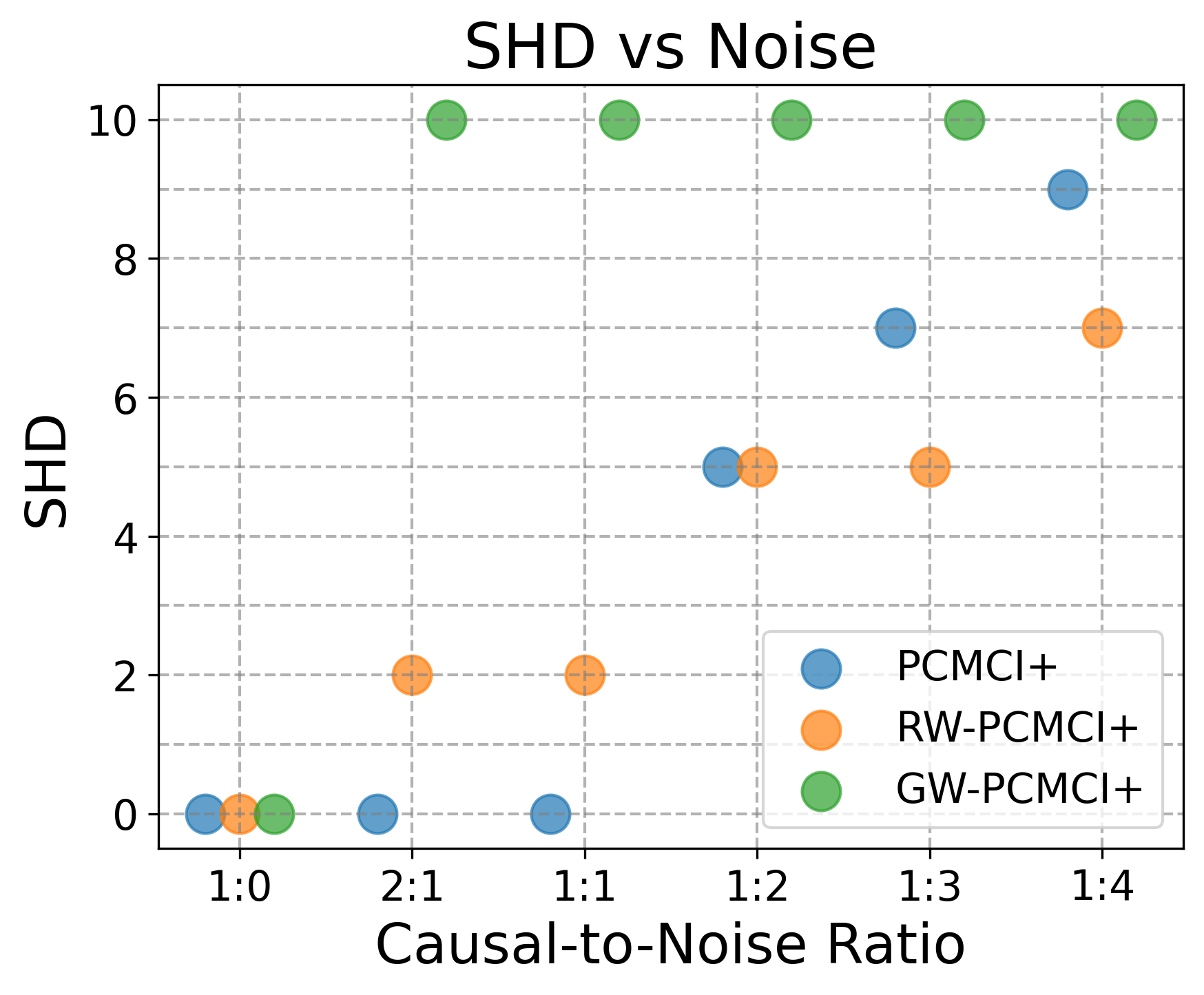}
    \end{subfigure}
    \centering
    \begin{subfigure}{0.24\columnwidth}
        \centering
        \includegraphics[width=\linewidth]{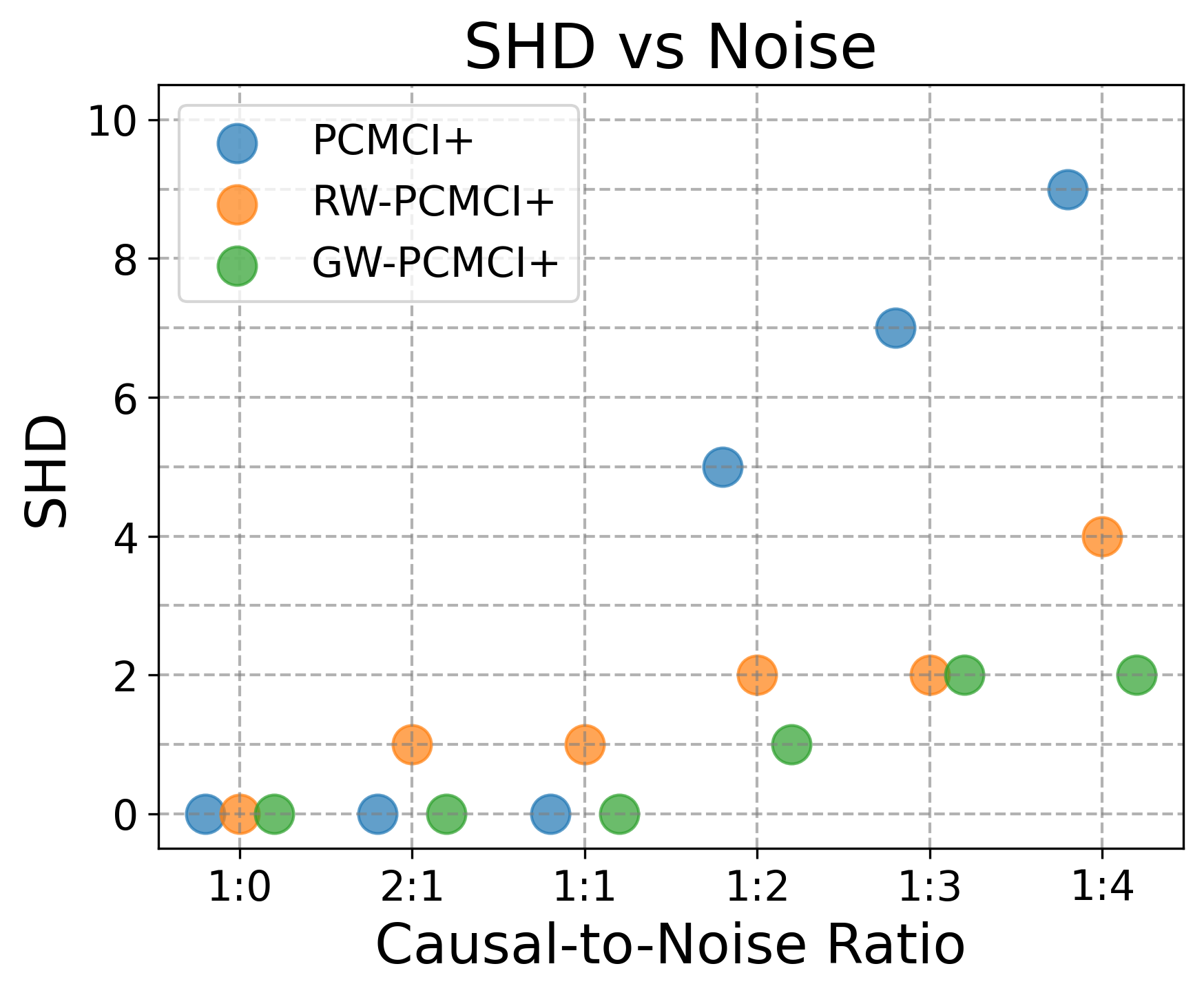}
    \end{subfigure}
    \hfill
    \begin{subfigure}{0.24\columnwidth}
        \centering
        \includegraphics[width=\linewidth]{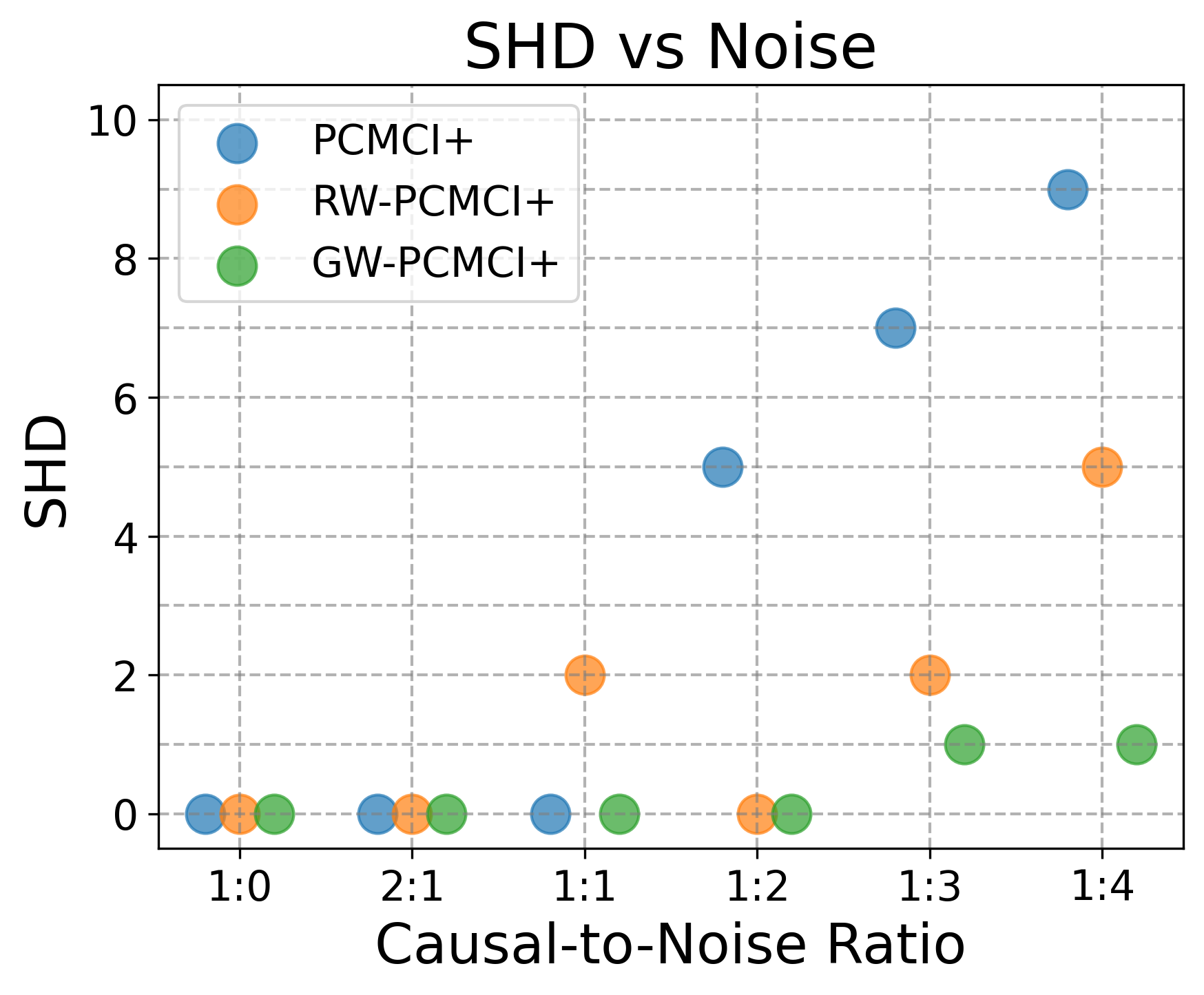}
    \end{subfigure}
    \centering
    \begin{subfigure}{0.24\columnwidth}
        \centering
        \includegraphics[width=\linewidth]{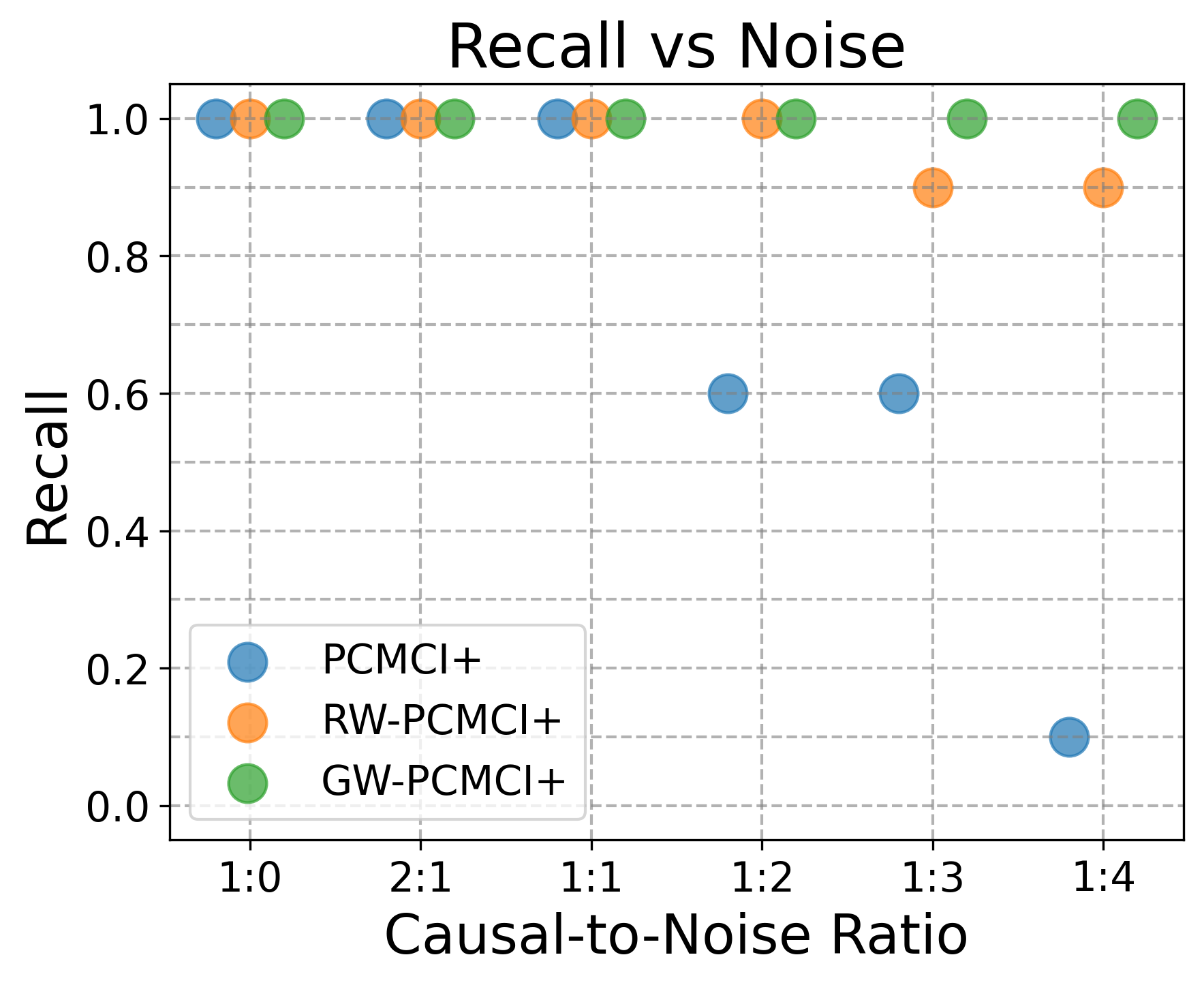}
    \end{subfigure}
    \hfill
    \begin{subfigure}{0.24\columnwidth}
        \centering
        \includegraphics[width=\linewidth]{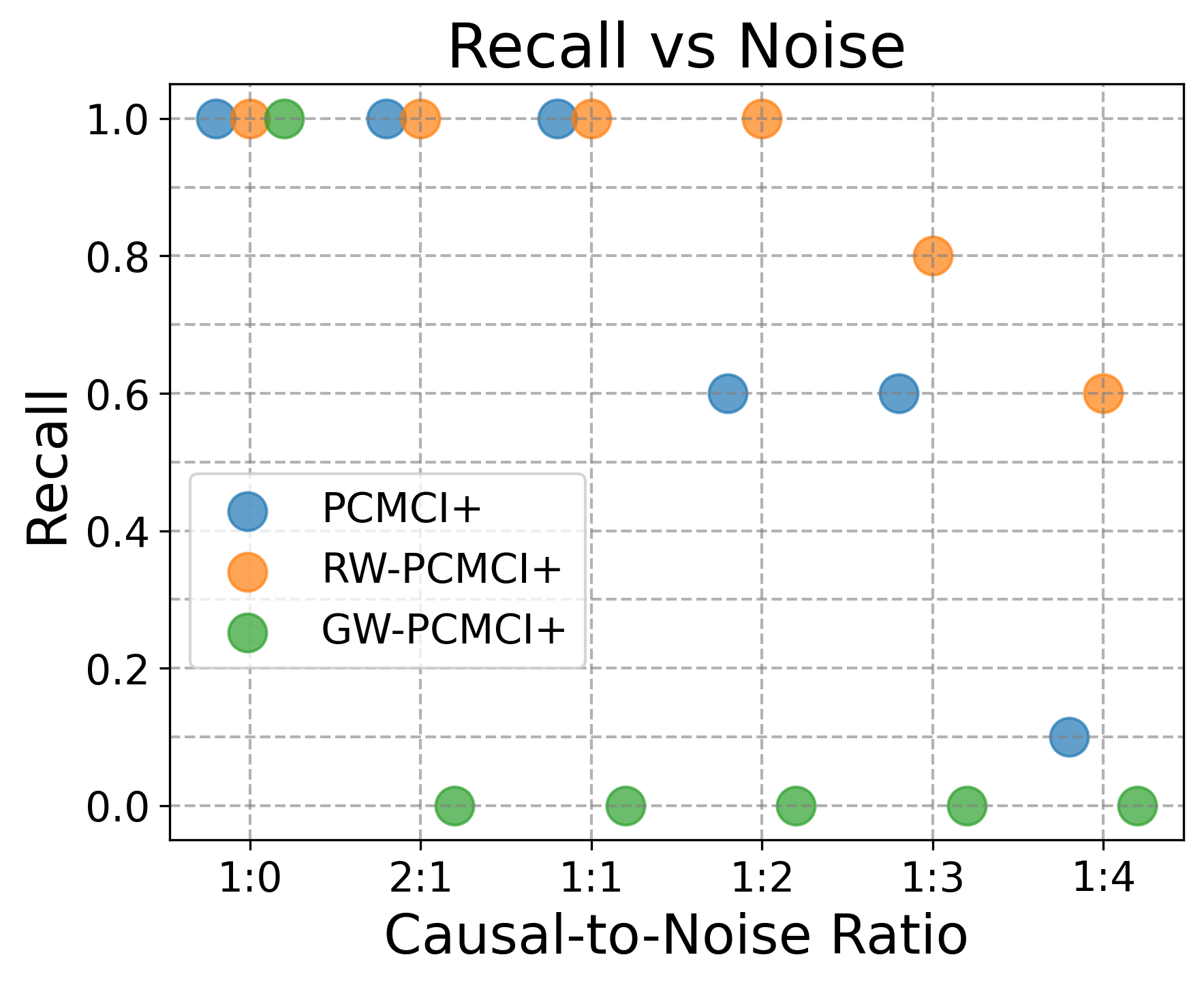}
    \end{subfigure}
    \centering
    \begin{subfigure}{0.24\columnwidth}
        \centering
        \includegraphics[width=\linewidth]{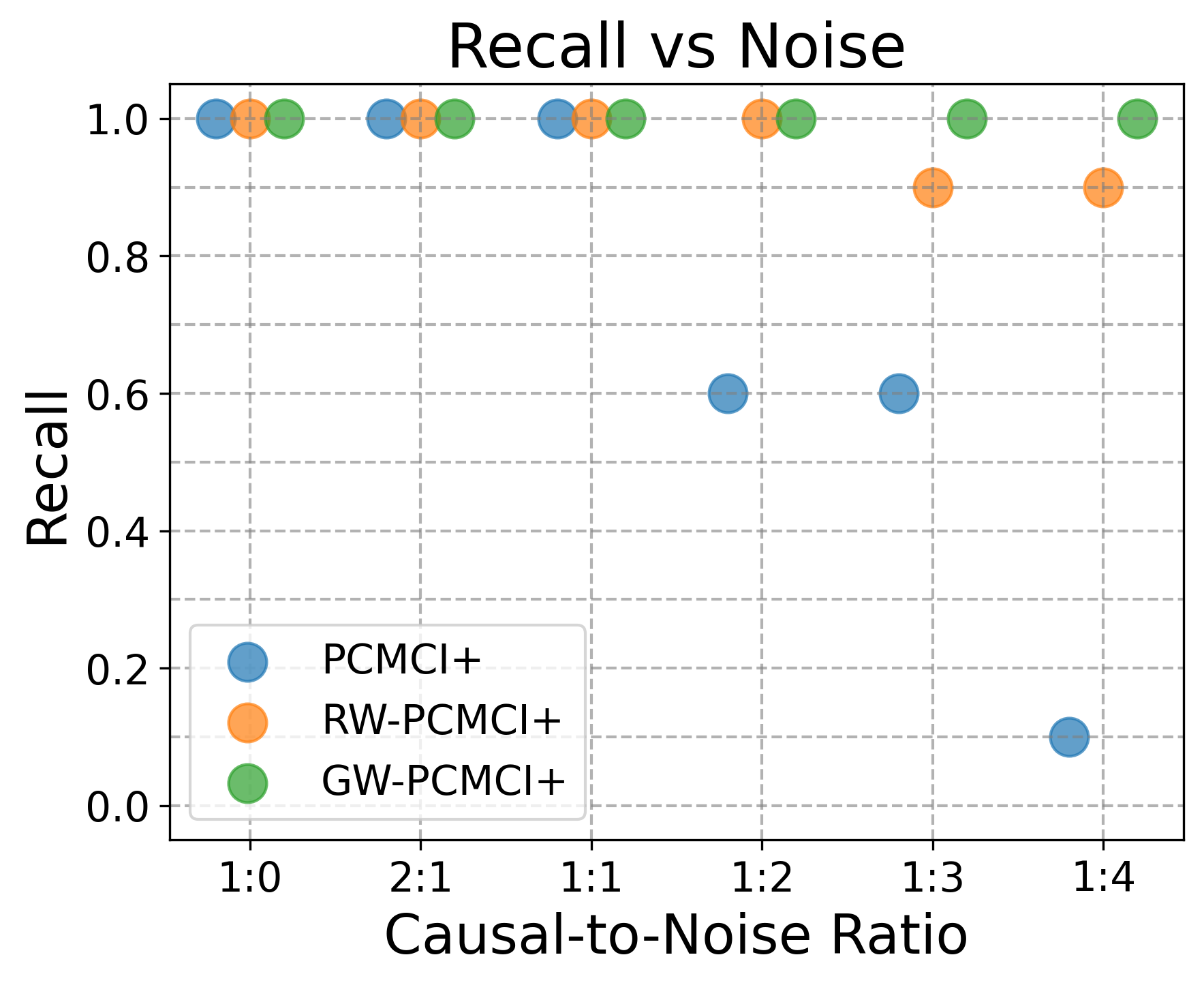}
    \end{subfigure}
    \hfill
    \begin{subfigure}{0.24\columnwidth}
        \centering
        \includegraphics[width=\linewidth]{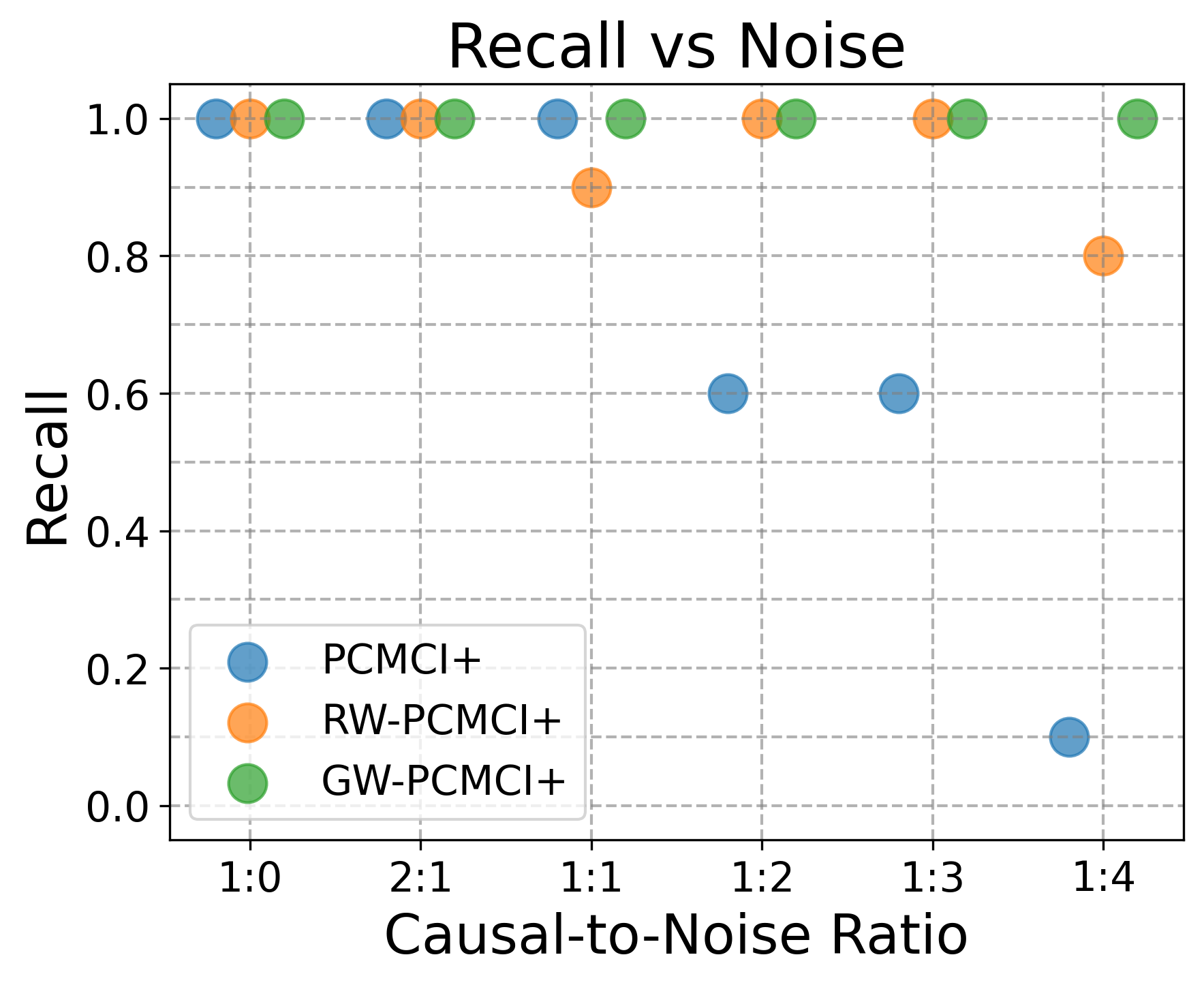}
    \end{subfigure}
       \centering
    \begin{subfigure}{0.24\columnwidth}
        \centering
        \includegraphics[width=\linewidth]{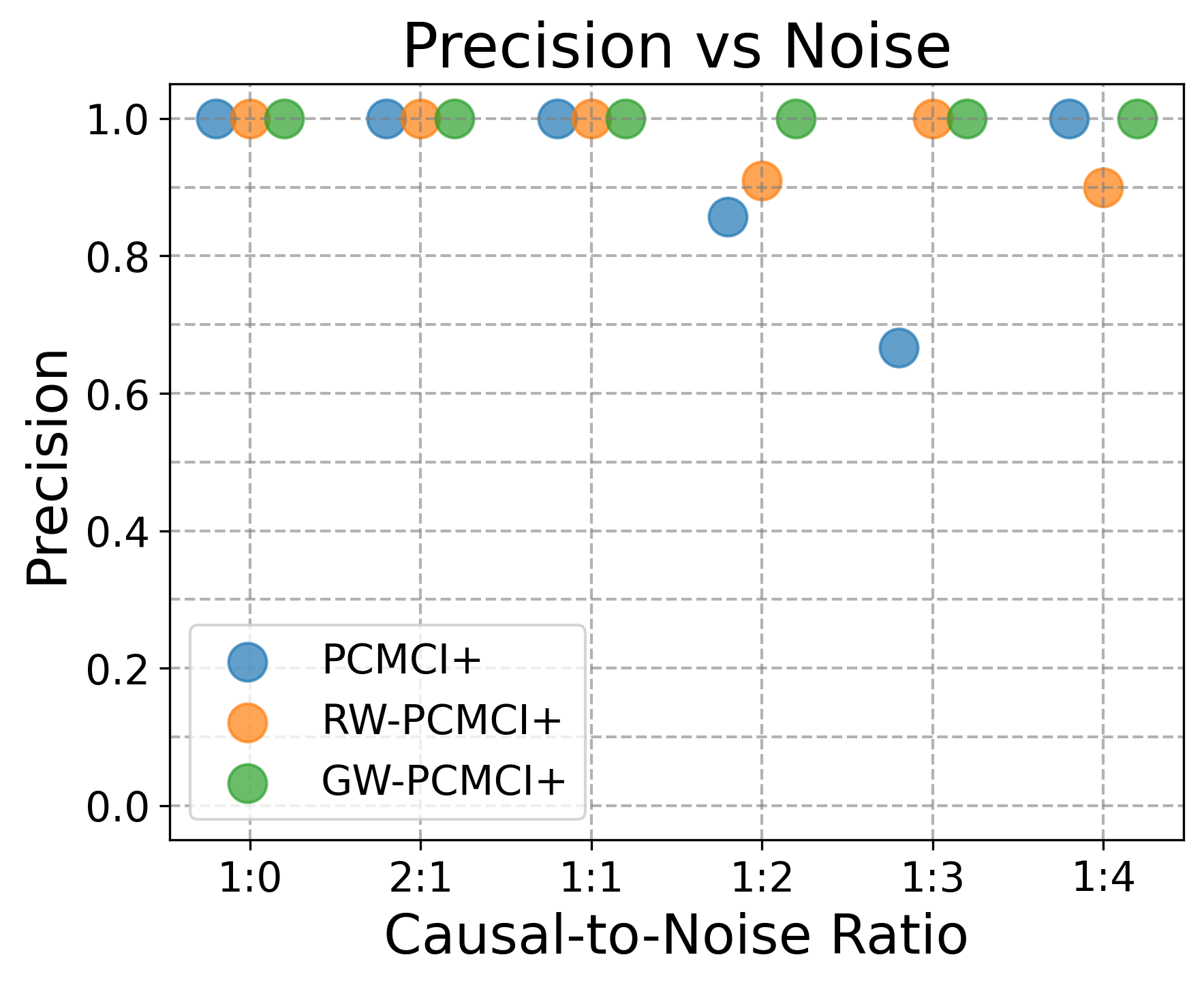}
    \end{subfigure}
    \hfill
    \begin{subfigure}{0.24\columnwidth}
        \centering
        \includegraphics[width=\linewidth]{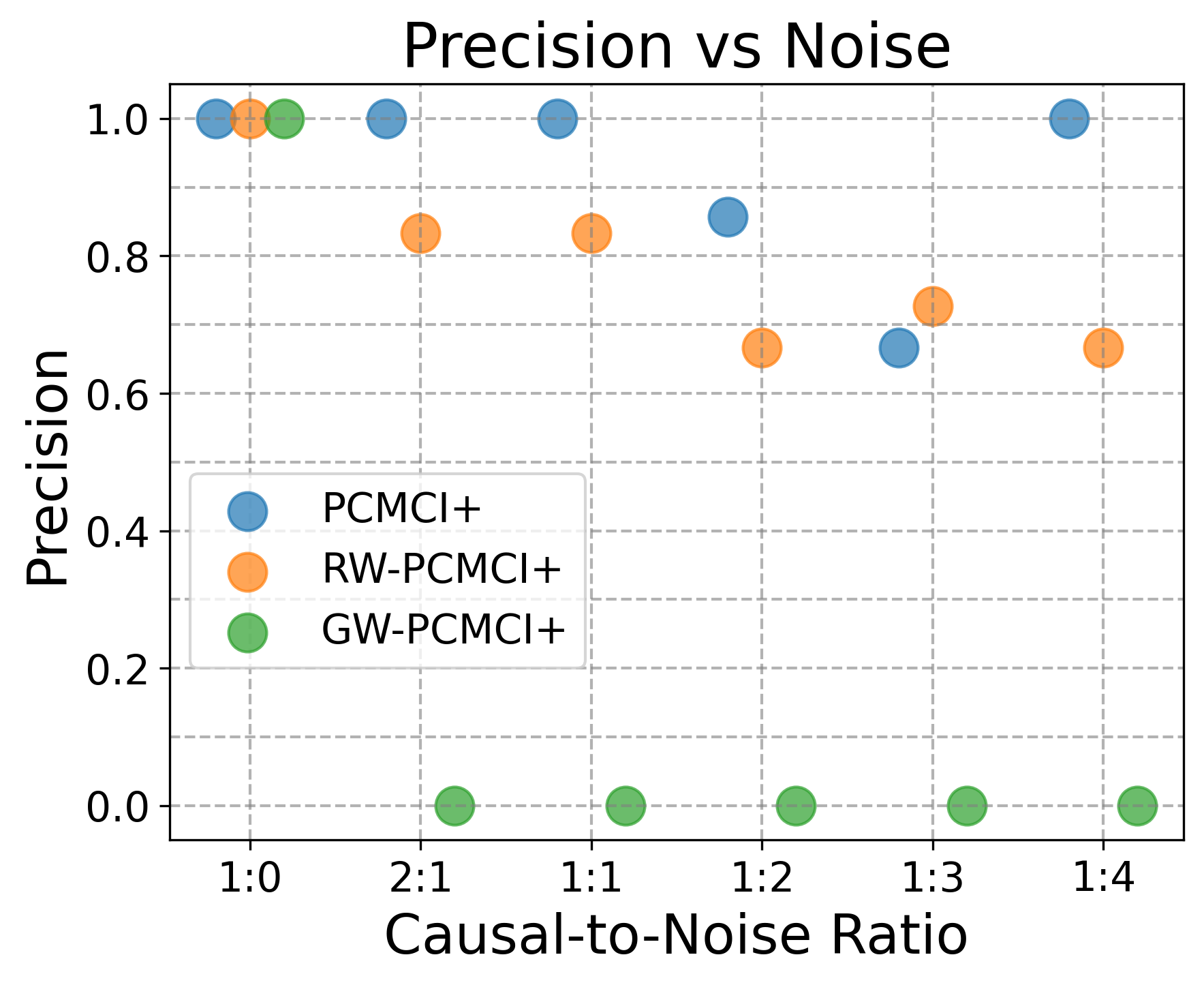}
    \end{subfigure}
           \centering
    \begin{subfigure}{0.24\columnwidth}
        \centering
        \includegraphics[width=\linewidth]{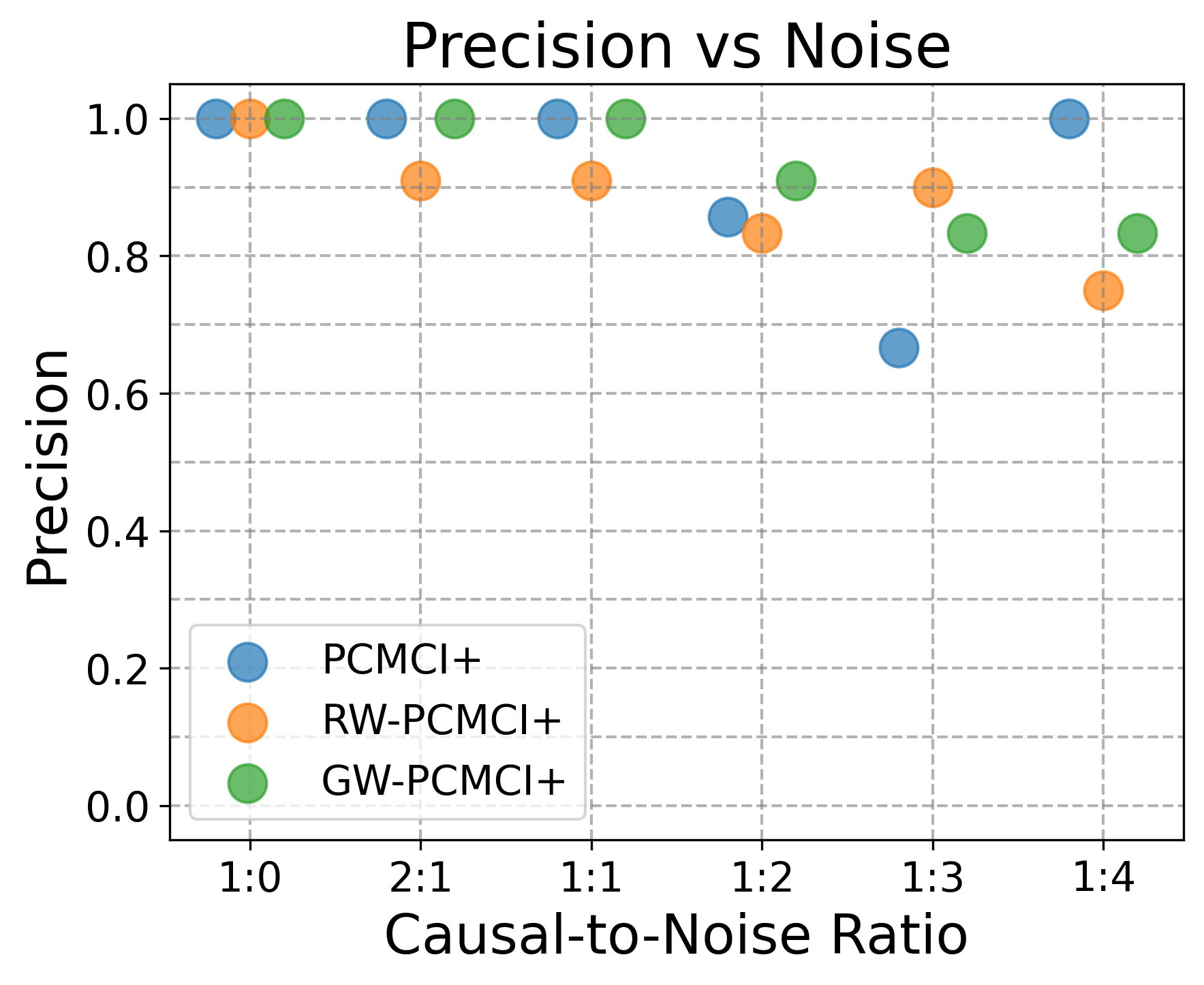}
    \end{subfigure}
    \hfill
    \begin{subfigure}{0.24\columnwidth}
        \centering
        \includegraphics[width=\linewidth]{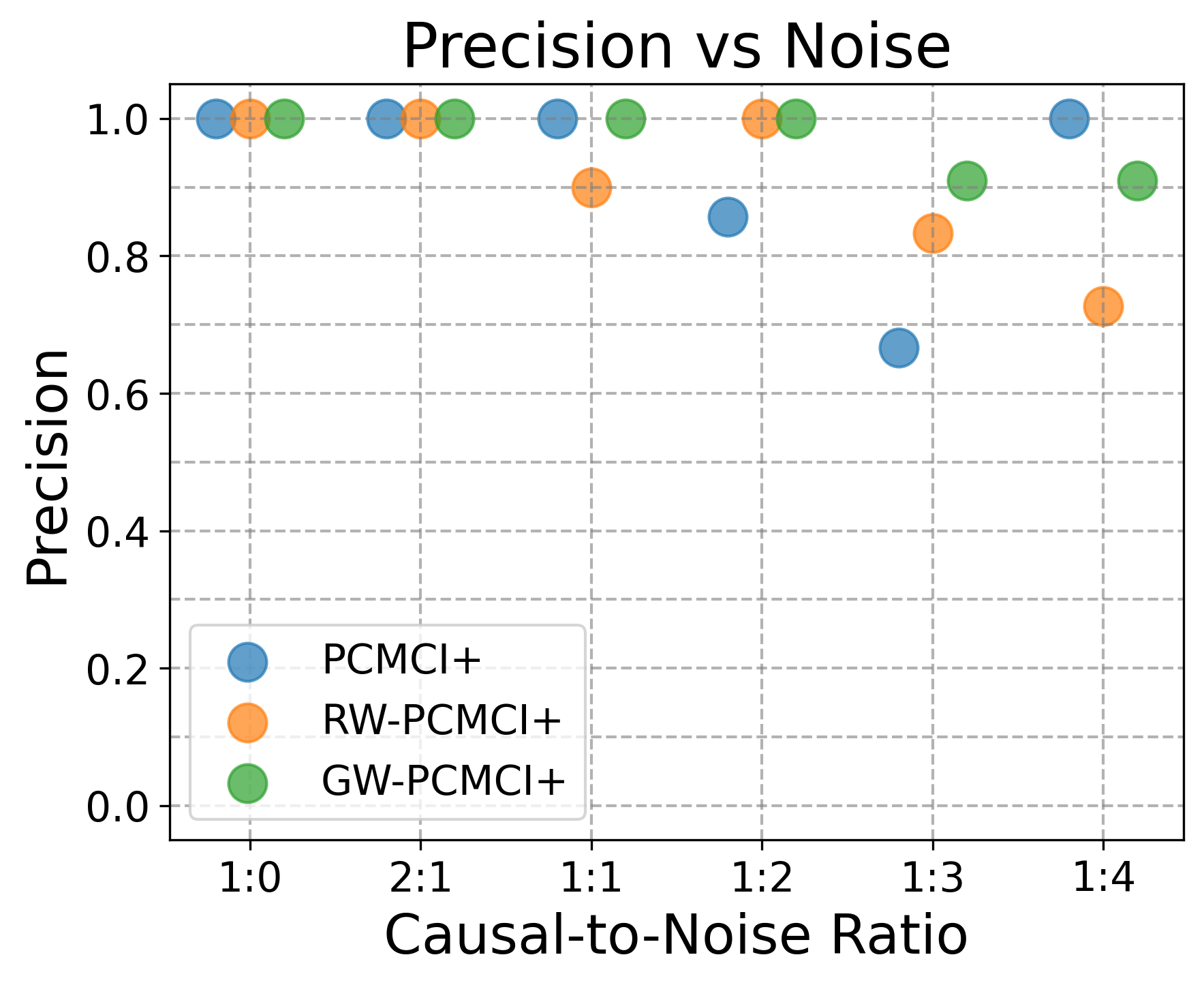}
    \end{subfigure}
         \caption{SHD, recall and precision for Experiments 1-4 (first through fourth columns, respectively) across varying causal-to-noise ratios, without considering contemporaneous edges. }
    \label{fig:results_lag0}
\end{figure}

When optimizing the p-value threshold that yields the best (i.e., lowest) SHD, both contemporaneous and lagged edges are taken into account. However, since our setting contains only lagged edges, it is also informative to analyse the results obtained when contemporaneous edges are excluded - that is, when only lagged edges are considered. These results are shown in Figure~\ref{fig:results_lag0}.
As can be seen, all methods generally exhibit improved performance under this evaluation, particularly in terms of precision. This suggests that nonexistent contemporaneous edges were often inferred with a higher certainty than some of the true lagged edges, thereby reducing overall performance. Investigating the interaction between contemporaneous and lagged edge estimation constitutes an important direction for future work.

\end{document}